\documentclass[runningheads]{llncs}

\usepackage[mobile]{eccv}
\usepackage{eccvabbrv}

\usepackage{graphicx}
\usepackage{booktabs}
\usepackage{paralist} 
\usepackage[accsupp]{axessibility}  

\usepackage[pagebackref,breaklinks,colorlinks]{hyperref}

\usepackage{orcidlink}
\usepackage{xcolor}
\usepackage{graphicx}
\usepackage{amsmath}
\usepackage{amssymb}
\usepackage{booktabs}
\usepackage{multirow}
\usepackage{bbm}
\usepackage{float}
\usepackage{graphicx}
\usepackage{amsmath}
\usepackage{amssymb}
\usepackage{booktabs}
\usepackage{multibib}
\usepackage{caption}
\usepackage{subcaption}
\usepackage{multirow}
\usepackage{url}            %
\usepackage{array}
\usepackage{graphicx}
\definecolor{mycolor}{HTML}{D8ECD1} 
\newcommand{\better}[1]{\colorbox{mycolor}{#1}}

\newlength\savewidth

\newcommand{\tablestyle}[2]{\setlength{\tabcolsep}{#1}\renewcommand{\arraystretch}{#2}\centering\footnotesize}

\newcommand{\datatag}[1]{\rotatebox[origin=l]{90}{\small{#1}}}

\makeatletter\renewcommand\paragraph{\@startsection{paragraph}{4}{\z@}
  {.5em \@plus1ex \@minus.2ex}{-.5em}{\normalfont\normalsize\bfseries}}\makeatother

\newcolumntype{x}[1]{>{\centering\arraybackslash}p{#1pt}}
\newcolumntype{y}[1]{>{\raggedright\arraybackslash}p{#1pt}}
\newcolumntype{z}[1]{>{\raggedleft\arraybackslash}p{#1pt}}

\newcommand{\app}{\raise.17ex\hbox{$\scriptstyle\sim$}}

\definecolor{baselinecolor}{gray}{.9}

\newcolumntype{*}{>{\global\let\currentrowstyle\relax}}
\newcolumntype{^}{>{\currentrowstyle}}
\newcommand{\rowstyle}[1]{\gdef\currentrowstyle{#1}#1\ignorespaces}

\definecolor{dt}{gray}{0.7}  %

\newcommand{\huggingface}{\raisebox{-1.5pt}{\includegraphics[height=1.05em]{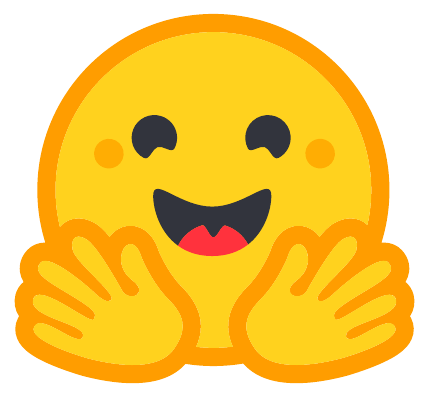}}\xspace}
\newcommand{\github}{\raisebox{-1.5pt}{\includegraphics[height=1.05em]{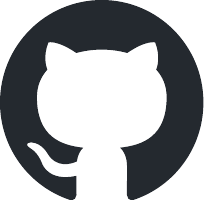}}\xspace}

\begin{document}

\title{Multi-label Cluster Discrimination \\ for Visual Representation Learning} 

\titlerunning{Multi-label Cluster Discrimination for Visual Representation Learning}

\author{Xiang An\inst{1}\orcidlink{0009-0008-4652-8296}
\and
Kaicheng Yang \inst{1}\orcidlink{0009-0008-6073-9014}
\and
Xiangzi Dai \inst{1}\orcidlink{0009-0009-3236-8380}
\and \\
Ziyong Feng \inst{1}\orcidlink{0009-0007-8689-8366}
\and
Jiankang Deng\thanks{Corresponding author.}\inst{2}\orcidlink{0000-0002-3709-6216}
}

\authorrunning{Xiang An, Kaicheng Yang, Xiangzi Dai, Ziyong Feng and Jiankang Deng}

\institute{
DeepGlint \\
\email{xiangan@deepglint.com} \and
Huawei Noah’s Ark Lab \\
\email{jiankang.deng@gmail.com}
}

\maketitle

\begin{abstract}
\label{sec:abstract}
Contrastive Language Image Pre-training (CLIP) has recently demonstrated success across various tasks due to superior feature representation empowered by image-text contrastive learning. However, the instance discrimination method used by CLIP can hardly encode the semantic structure of training data. To handle this limitation, cluster discrimination has been proposed through iterative cluster assignment and classification. Nevertheless, most cluster discrimination approaches only define a single pseudo-label for each image, neglecting multi-label signals in the image.
In this paper, we propose a novel Multi-Label Cluster Discrimination method named MLCD to enhance representation learning. In the clustering step, we first cluster the large-scale LAION-400M dataset into one million centers based on off-the-shelf embedding features. Considering that natural images frequently contain multiple visual objects or attributes, we select the multiple closest centers as auxiliary class labels. In the discrimination step, we design a novel multi-label classification loss, which elegantly separates losses from positive classes and negative classes, and alleviates ambiguity on decision boundary. We validate the proposed multi-label cluster discrimination method with experiments on different scales of models and pre-training datasets. Experimental results show that our method achieves state-of-the-art performance on multiple downstream tasks including linear probe, zero-shot classification, and image-text retrieval. Code and models have been released at \github 
\url{https://github.com/deepglint/unicom} and \huggingface \href{https://huggingface.co/collections/DeepGlint-AI/mlcd-670d18d767cea37ea7436e69}{Hugging Face}.


\end{abstract}

\section{Introduction}

\label{sec:intro}

\begin{figure}[t]
\centering
\begin{subfigure}{0.33\textwidth}
\includegraphics[height=1.3\textwidth]{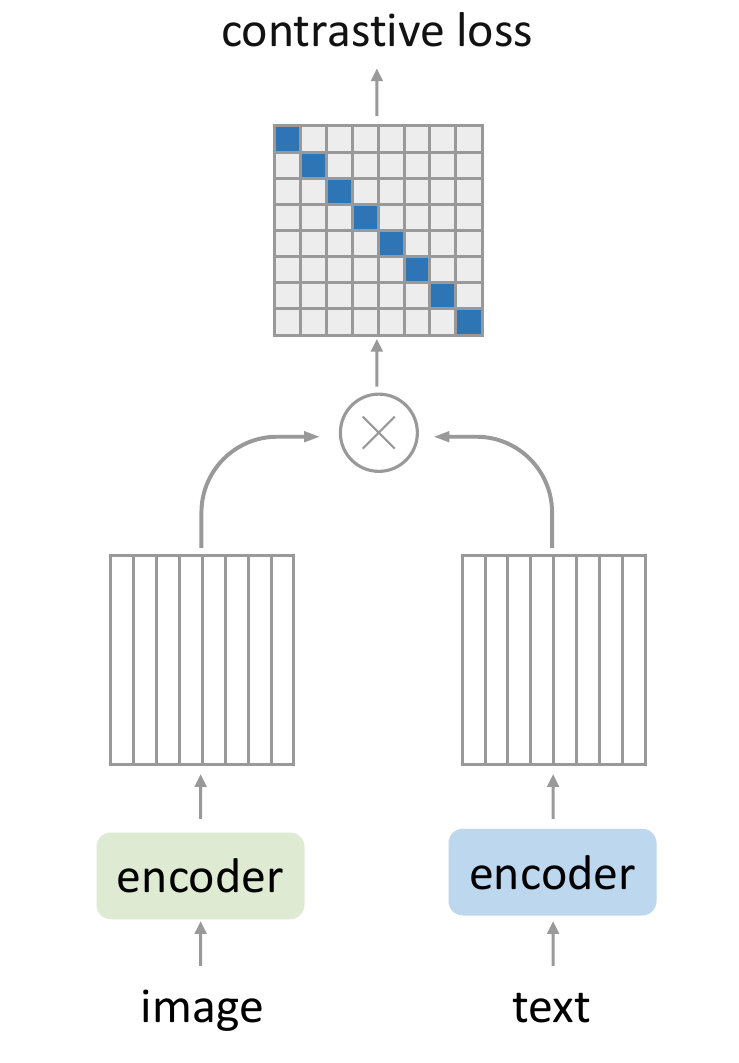}
\captionsetup{justification=centering}
\caption{ Instance \linebreak  Discrimination}
\label{fig1:instance}
\end{subfigure}\hspace{-10pt}
\begin{subfigure}{0.33\textwidth}
\includegraphics[height=1.3\textwidth]{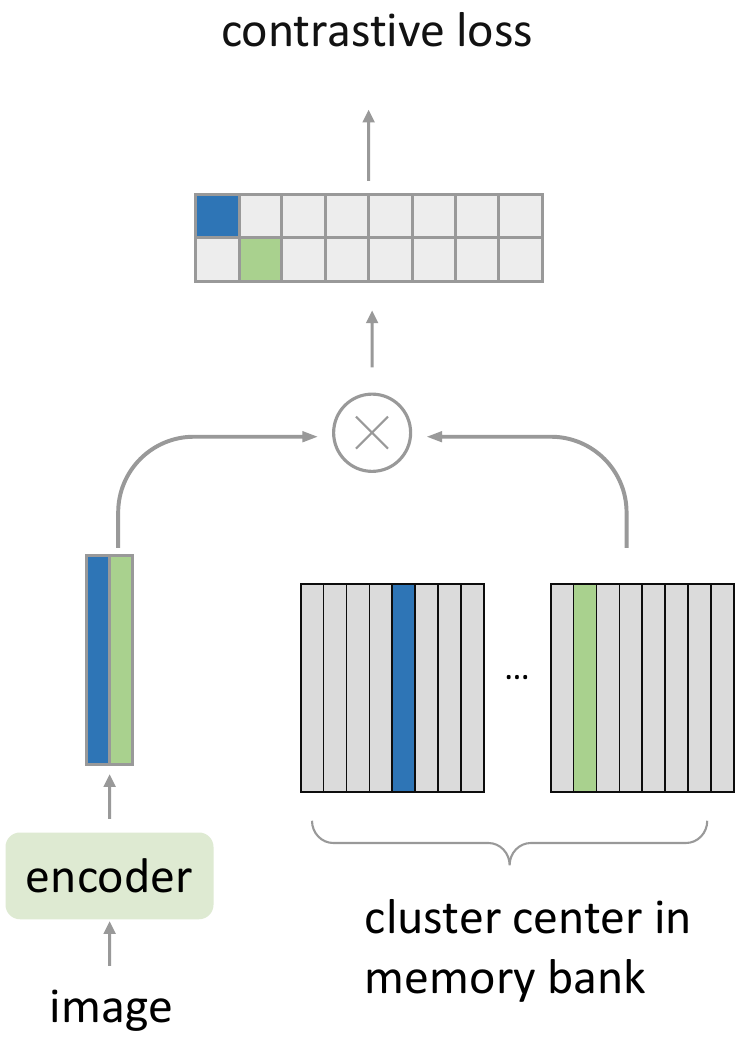}
\captionsetup{justification=centering}
\caption{ Cluster \linebreak Discrimination}
\label{fig1:cluster}
\end{subfigure}\hspace{-5pt}
\begin{subfigure}{0.33\textwidth}
\includegraphics[height=1.3\textwidth]{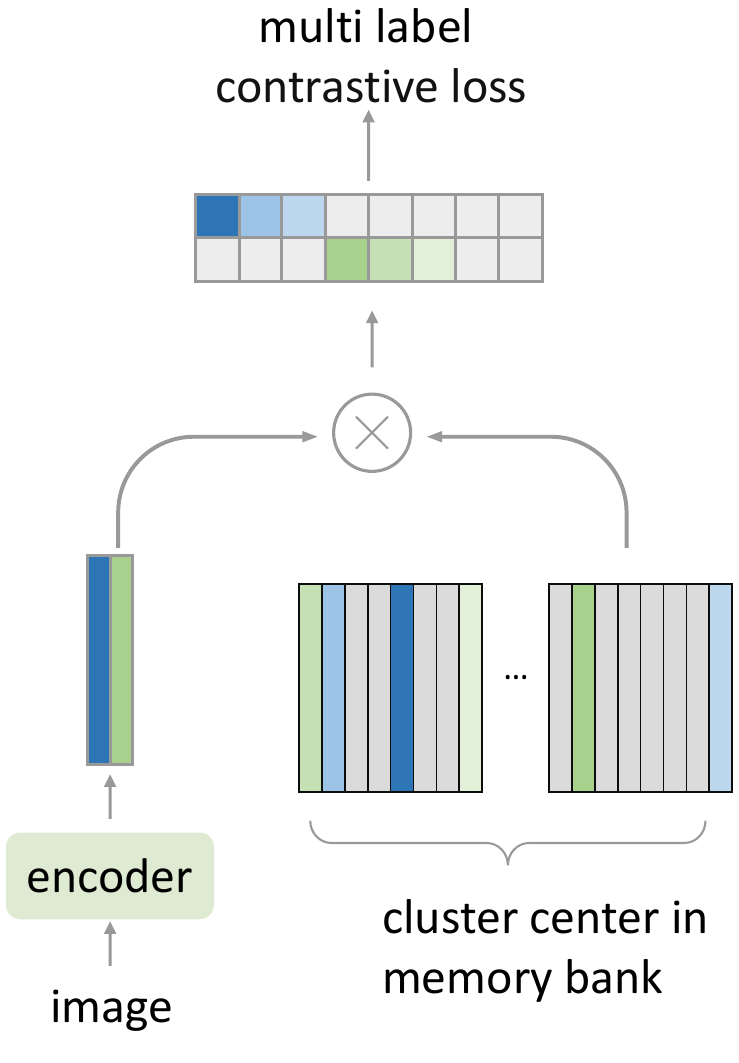}
\captionsetup{justification=centering}\caption{ Multi-Label \linebreak Cluster Discrimination}
\label{fig1:multicluster}
\end{subfigure}
\caption{Comparisons of instance discrimination, cluster discrimination, and the proposed multi-label cluster discrimination. (a) Instance discrimination treats each image-text pair as a unique instance, failing to capture the semantic structure within the training data. (b) Cluster discrimination improves the semantic embedding by grouping similar instances but struggles with multi-label signals in a single image. (c) The proposed multi-label cluster discrimination addresses this challenge by assigning multiple class labels to each sample, 
capturing different granularities of visual signals (\eg, objects or attributes) in one image.}
\label{fig1:main}
\vspace{-4mm}
\end{figure}

Language-supervised visual pre-training, \eg, CLIP~\cite{radford2021learning} and ALIGN~\cite{jia2021scaling}, 
has been established as a simple yet effective methodology for visual representation learning. Empowered by image-text contrastive learning, pre-trained CLIP models exhibit remarkable versatility and transferability across various downstream tasks (\eg, linear probe, zero-shot classification, and image retrieval). As illustrated in Fig.~\ref{fig1:instance}, CLIP aligns the visual and textual signals of each instance into a unified semantic space by cross-modal instance discrimination. Nevertheless, the instance discrimination method used by CLIP can hardly encode the semantic structure of training data, because instance-wise contrastive learning always treats two samples as a negative pair if they are from different instances, regardless of their semantic similarity. When a large number of instances are selected into the mini-batch to form the contrastive loss, negative pairs that share similar semantics will be undesirably pushed apart in the embedding space. 

To handle the limitations of instance discrimination, cluster discrimination methods (\eg, DeepCluster~\cite{caron2018deep}, SeLa~\cite{asano2019self}, ODC~\cite{zhan2020online}, SwAV~\cite{caron2020unsupervised}, CoKe~\cite{qian2022unsupervised}, and UNICOM~\cite{an2023unicom}) have been proposed for deep unsupervised learning through jointly learning image embeddings and cluster assignments. Learning representations with clusters will pull similar instances together, which is beneficial for capturing semantic structures in data. However, most cluster discrimination approaches only define a single pseudo-label for each image as depicted in Fig.~\ref{fig1:cluster}. By contrast, natural language supervision proposed in CLIP can provide richer forms of labels for a single image, \eg, objects, scenes, actions, and relations, at multiple levels of granularity. 

As can be seen from Fig.~\ref{fig:pruityconflict}, 
a web image frequently contains multiple classification targets, such as objects~\cite{yang2016exploit} or attributes~\cite{pham2021learning}. The existence of multiple objects in the image requires laborious cropping~\cite{Li_2023_ICCV,Abdelfattah_2023_ICCV} to construct single-label annotations, while some scenario elements and attributes in the image are hard to disentangle to obtain single-label instances~\cite{pham2021learning,Zhu_2023_ICCV}. These real-world challenges pose so-called multi-label classification where an image is equipped with multiple labels beyond a single label.

In this paper, we aim to boost the visual representation power of the CLIP model by introducing a novel Multi-Label Cluster Discrimination (MLCD) approach.
In the clustering step, we follow UNICOM~\cite{an2023unicom} to conduct one step of offline clustering by using the features predicted by a pre-trained CLIP model. Due to the limited discrimination power of the CLIP model~\cite{radford2021learning}, the single pseudo-label may not cover all of the visual signals (\eg, objects or attributes) in the image. To this end, we further perform a similarity-based sorting against $k$ class centers and select the top $l$ class centers as the positive class centers for that image. 
In the discrimination step, we follow the Circle loss~\cite{sun2020circle} to 
design a multi-label loss to effectively deal with multiple labels. The vanilla version of 
the multi-label loss exploits relative similarity comparisons between positive and negative classes. More specifically, the optimization seeks to narrow the gap between the intra-class similarities $\{s_i\}$ and 
the inter-class similarities $\{s_j\}$ by reducing all possible $(s_j-s_i)$. However, optimizing $(s_j-s_i)$
usually leads to a decision boundary allowing ambiguity~\cite{sun2020circle}. To this end, we introduce another two optimization targets (\ie, decreasing $s_j$ and increasing $s_i$) into the loss function. Introducing the additional two items enables an elegant separation of positive class loss and negative class loss (Eq.~\ref{eqn:mlc2}), {which can alleviate the ambiguity on the decision boundary}. To alleviate inter-class conflict and save the computation time on the classifier layer, we also employ PartialFC~\cite{an2022killing} and randomly sample part of the negative class centers during each iteration.  

The main contributions of our paper are the following:
\begin{enumerate}
\item We propose a novel multi-label cluster discrimination method for visual representation learning on large-scale data. In the clustering step, we employ one step of offline k-means to predict multiple labels for each training sample. In the discrimination step, we explore multi-label classification, which considers multiple supervision signals for a single image and learns better semantic structure in data.
\item To avoid ambiguity during the optimization of $(s_j-s_i)$, we add additional optimization targets by maximizing the within-class similarity $s_i$, as well as to minimizing the between-class similarity $s_j$. By doing so, the loss from positive class labels and negative class labels can be elegantly separated.
\item The proposed multi-label cluster discrimination significantly boosts the representation power compared to the instance discrimination-based model (\eg, OpenCLIP~\cite{cherti2023reproducible} and FLIP~\cite{li2023scaling}) and the cluster discrimination-based model (\eg, UNICOM~\cite{an2023unicom}) on the downstream tasks (\eg, linear probe, zero-shot classification, zero-shot retrieval).
\end{enumerate}

\section{Related Work}

\noindent{\bf Visual Representation Learning.}
Visual representation pre-training methods can be mainly divided into three categories: (1) supervised learning by using manually annotated class labels (\eg, ImageNet-1K/-21K~\cite{deng2009imagenet} and JFT-300M/-3B~\cite{dosovitskiy2021image,zhai2022scaling}), (2) weakly-supervised learning by employing hashtags~\cite{mahajan2018exploring,singh2022revisiting} or text descriptions~\cite{radford2021learning,jia2021scaling,li2023scaling}, and (3) unsupervised learning~\cite{chen2020big,he2020momentum,caron2018deep} by 
designing appropriate pretext tasks (\eg, solving jigsaw puzzles~\cite{noroozi2016unsupervised}, invariant mapping~\cite{chen2021exploring}, and masked image inpainting~\cite{he2022masked}). Even though fully supervised pre-training can learn a strong semantic signal from each training example, manual label annotation is time-consuming and expensive thus supervised learning is less scalable. In this paper, we focus on annotation-free pre-training which can be easily scaled to billions of web images to learn visual representation for downstream tasks. 

\noindent{\bf Instance and Cluster Discrimination.}
Instance discrimination~\cite{chen2020big,he2020momentum,radford2021learning} is usually implemented by the contrastive loss to pull images from the same instance as well as push away images from different instances. Among these instance discrimination methods,  
language-supervised visual pre-training, \eg, CLIP~\cite{radford2021learning,yang2023alip,gu2024rwkv},
is a simple yet powerful approach to take advantage of rich forms of labels at multiple levels of granularity for a single image. Even though CLIP~\cite{radford2021learning} has recently demonstrated impressive success, instance-wise contrastive learning always treats different instances as negative pairs thus it can hardly capture the full semantic information from the training data. 

To explore potential semantic structures in the training data, cluster discrimination~\cite{caron2018deep,asano2019self,zhan2020online,li2020prototypical,caron2020unsupervised,qian2022unsupervised} is proposed with two iterative steps: (1) the clustering step to assign a single class label for each sample, and (2) the classification step to learn a classifier to predict the assigned pseudo label. In cluster discrimination methods, each cluster contains more than one instance, visually similar instances will be pulled closer and thus cluster discrimination can better capture semantic structures from data. However, multiple visual elements can exist in one single image and the single label used by cluster discrimination may not cover all visual signals. 

\noindent{\bf Multi-label Classification.} 
Multi-label classification~\cite{tsoumakas2007multi,zhang2013review} assigns a set of multiple labels for each instance. Compared with single-class classification, where each instance is assigned with a single label, multi-label classification~\cite{yang2016exploit,zhao2021transformer,Xia_2023_ICCV} is more challenging~\cite{liu2017easy,liu2021emerging}. Considering multiple labels are drawn from $k$ categories, the multi-label classification can be decomposed into $k$ binary classification tasks. However, the binary cross-entropy loss involves issues regarding imbalance~\cite{ridnik2021asymmetric}. Through analyzing the intrinsic loss functions
of the classification loss and the metric loss~\cite{wang2019multi}, Sun \etal~\cite{sun2020circle} formulate a unified multi-label loss function to exploit relative comparison between positive and negative classes. Nevertheless, the relative comparison $(s_j-s_i)$ allows ambiguity for convergence. Su \etal~\cite{su2022zlpr} introduce a threshold into the multi-label loss and design the Threshold-bounded Log-sum-exp and Pairwise Rank-based (TLPR) loss, hoping that the logits of positive categories will be larger than the threshold and the logits of negative categories will be smaller than the threshold. However, the TLPR loss is only designed for clean multi-label datasets and is not suitable for large-scale multi-label datasets with heavy noises.
In this paper, we only employ one step of offline clustering to predict multiple labels for each image and then design a robust multi-label classification disambiguation loss to achieve good feature representation when training on the automatically clustered large-scale data.

\section{Method}
Given a training set $X=\{x_1, x_2,...,x_n\}$ including $n$ images, visual representation learning aims at learning a function $f$ that maps images $X$ to normalized embeddings $E=\{e_1, e_2,...,e_n\}$ with $e_i=f(x_i)$, such that embeddings can describe the semantic similarities between different images. 

\subsection{Preliminaries}

\noindent\textbf{Instance Discrimination} achieves semantic embedding by minimizing a contrastive loss function represented as:
\vspace{-2mm}
\begin{equation}
\label{eqn:ID}
\mathcal{L}_\mathrm{ID} = - \log \frac{\exp(e_i'^T e_i   )}{\sum_{j=1}^k\exp(e_j'^T e_i   ) },
\end{equation}
where $\exp(\cdot)$ denotes the exponential function, and $e_i$ and $e_i'$ denote the normalized image and text embeddings for the instance $i$ in CLIP~\cite{radford2021learning}. Meanwhile, $e_j'$ contains one positive text representation for $i$ and $(k-1)$ negative text representations sourced from different instances. As illustrated in Fig.~\ref{fig1:instance}, the instance discrimination based CLIP model jointly trains an image encoder and a text encoder to predict the correct image-text pairings from a batch of training examples.

\noindent{\bf Cluster Discrimination}
is composed of two primary stages: the clustering process and the discrimination process.
During the clustering phase, every instance is assigned one pseudo-class label. This label is later employed as a guiding factor for training a classifier in the subsequent discrimination phase. 
For the normalized embedding feature $e_i = f(x_i)$,
the clustering process determines a centroid matrix $W \in \mathbb{R}^{d\times k}$ and assigns the cluster label $y_i$ for each image $x_i$. This is achieved by
\vspace{-2mm}
\begin{equation}
\label{eq:kmeans}
\min_{W \in \mathbb{R}^{d\times k}}
\frac{1}{n}
\sum_{i=1}^n
\min_{y_i \in \{0,1\}^{k}}
\| e_i  -  W y_i \|_2^2
\quad
\text{s.t.}
\quad
y_i^\top \bf{1}_k = 1,
\end{equation}
where $n$ is the number of training samples, $e_i$ is the normalized feature embedding obtained by using the image encoder $f$, and
the centroid $w_i$ belonging to centroid matrix $W\in {R}^{d\times k}$ is considered the normalized prototype of $i$-th cluster. $y_i$, falling within the set $\{0,1\}^k$, stands as a single label assignment restricted by the condition $y_i^\top \bf{1}_k = 1$, where $ \bf{1}_k $ is 1-vector with a length of $k$. 

Then, the training data, denoted as $\{x_i\}_{i=1}^n$, is divided into $k$ classes represented by prototypes $W=\{w_i\}_{i=1}^k$. 
Utilizing the pseudo labels and centroids derived from the clustering phase, the process of cluster discrimination can be executed by minimizing a conventional softmax classification loss, formulated as:
\vspace{-2mm}
\begin{align}
\label{eqn:cd}
 \mathcal{L}_\mathrm{CD} 
& = - \log \frac{\exp( w_{y_i}^T e_i)}{\sum_{j=1}^{k} \exp(w_j^Te_i)} 
   = - \log \frac{\exp({s_i})}{\sum_{j=1}^{k} \exp({s_j})} \notag \\ 
&  = \log (1 + \sum_{j=1, j \neq i}^{k} \exp(s_j-s_i)), 
\end{align}
where $e_i$ is the normalized embedding corresponding to the image $x_i$, and $x_i$ is categorized under the class symbolized by the normalized prototype $w_{y_i}$. For a more straightforward representation, we define the intra-class similarity $w_{y_i}^Te_i$, and the inter-class similarity, $w_j^Te_i$ as $s_i$ and $s_j$, respectively. 
Based on Eq.~\ref{eqn:cd}, in the discrimination phase that employs classification, $s_j$ and $s_i$ are paired to optimize the reduction of the difference ($s_j-s_i$). As depicted in Fig.~\ref{fig1:cluster}, the cluster discrimination based UNICOM model~\cite{an2023unicom} trains an image encoder to predict the one-hot pseudo label for each image from a batch of training examples.

\definecolor{color1}{RGB}{182,0,76}
\definecolor{color2}{RGB}{0,106,91}

\begin{figure}[!t]
\centering
\includegraphics[width=0.95\textwidth]{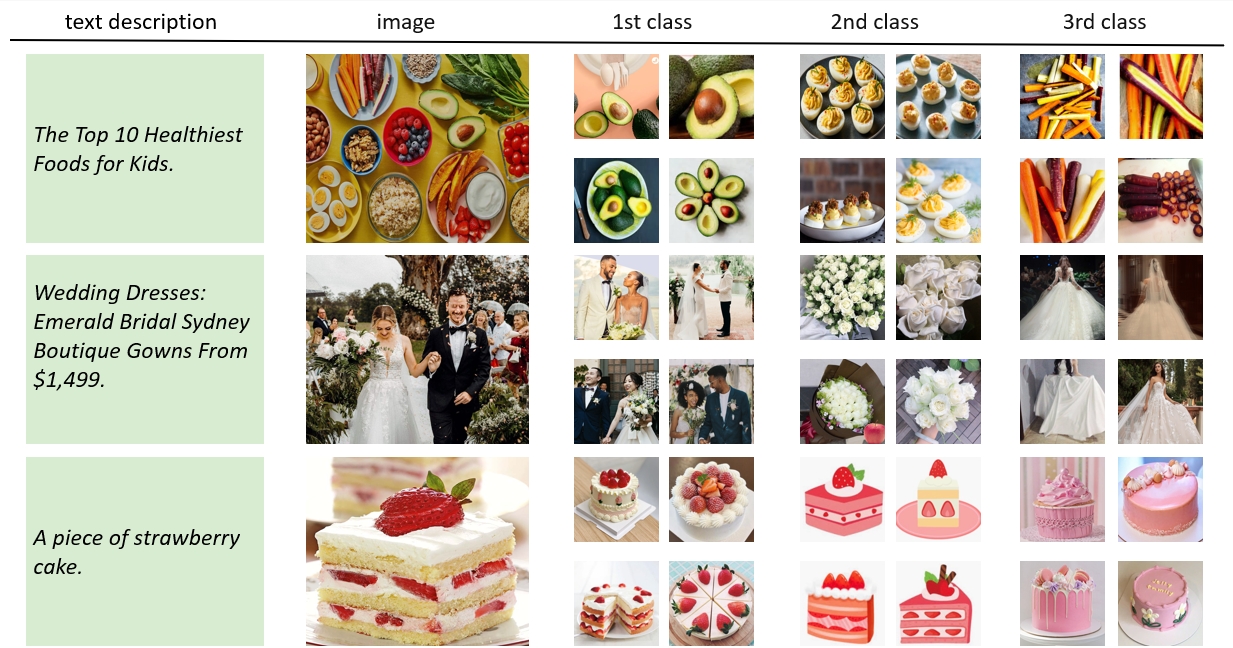}
\caption{Illustration of the multiple visual elements (\eg, objects or attributes) in images from the automatically clustered LAION-400M dataset.}
\label{fig:pruityconflict}
\vspace{-4mm}
\end{figure}

\subsection{Multi-label Cluster Discrimination} 

\noindent\textbf{Clustering.}
Considering the time consumption of iterative clustering and discrimination~\cite{caron2018deep}, An \etal~\cite{an2023unicom} implemented a single step of offline clustering with the aid of the pre-trained CLIP model (\ie, ViT-L/14) and efficient feature quantization~\cite{johnson2019billion}. On the large-scale LAION-400M dataset, it only takes around 10 minutes to cluster one million classes. Despite the straightforwardness of the clustering step, the automatically clustered large-scale dataset inevitably confronts intra-class purity and inter-class conflict problems due to the specific definition of class granularity.

In the realm of clustering algorithms, there often exists a trade-off between maintaining high within-class purity and ensuring low inter-class conflict. In the context of contrastive learning, the issue of inter-class conflict can be significantly alleviated by reducing the number of sampled negative instances within the mini-batch and adopting a suitable semi-hard mining technique. In this paper, we follow UNICOM~\cite{an2023unicom} to prioritize intra-class purity (\ie, clustering one million level classes from 400 million images) and employ margin-based PatialFC~\cite{an2022killing,deng2019arcface} to alleviate inter-class conflict (\ie, randomly sampling part of the negative class centers during each iteration). 

\begin{figure}[!t]
\centering
  \begin{subfigure}{0.24\textwidth}
    \centering
    \includegraphics[width=\textwidth]{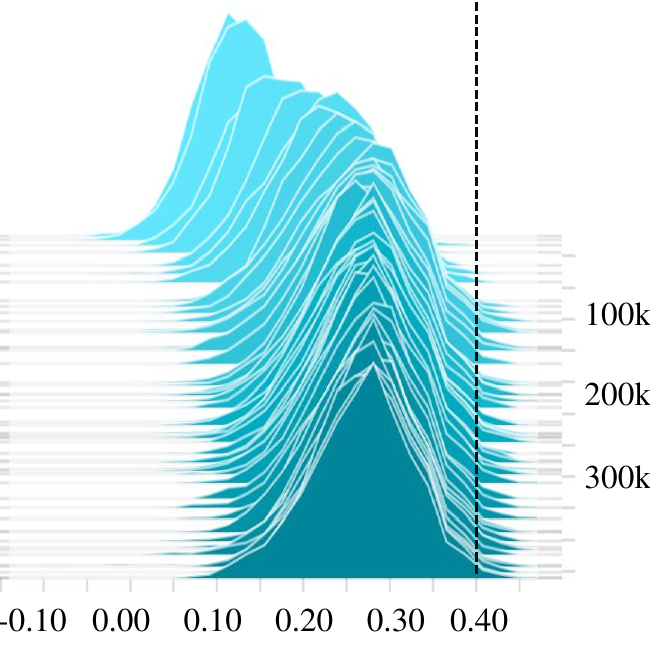}
    \caption{$s_i$ of MLC} 
    \label{cosine:mlc}
  \end{subfigure}
  \begin{subfigure}{0.24\textwidth}
    \centering
    \includegraphics[width=\textwidth]{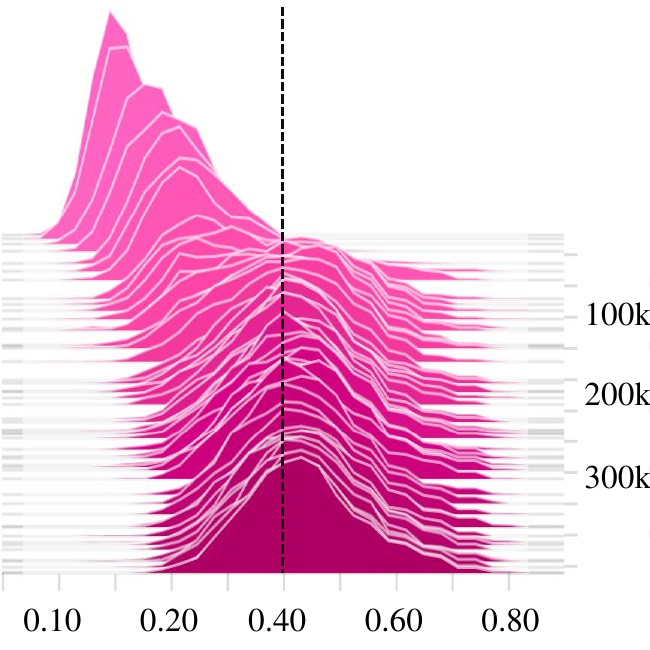}
    \caption{$s_i$ of MLCD}
    \label{cosine:mlcd}
  \end{subfigure}
  \begin{subfigure}{0.24\textwidth}
    \centering
    \includegraphics[width=\textwidth]{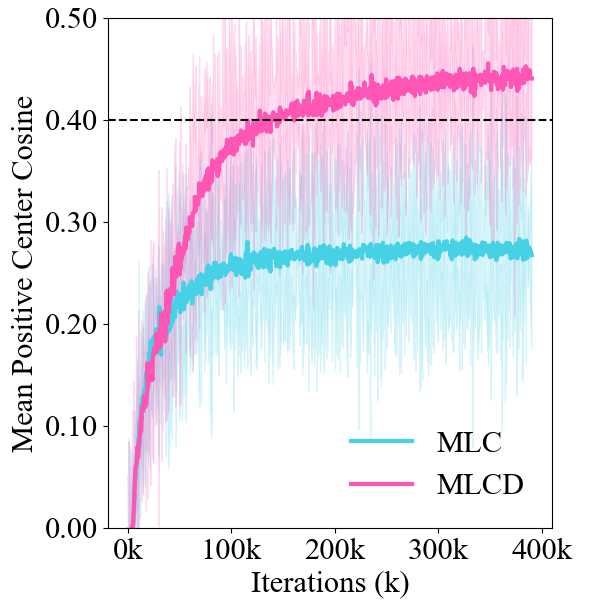}
    \caption{Mean $s_i$}
    \label{cosine:si}
  \end{subfigure}
  \begin{subfigure}{0.24\textwidth}
    \centering
    \includegraphics[width=\textwidth]{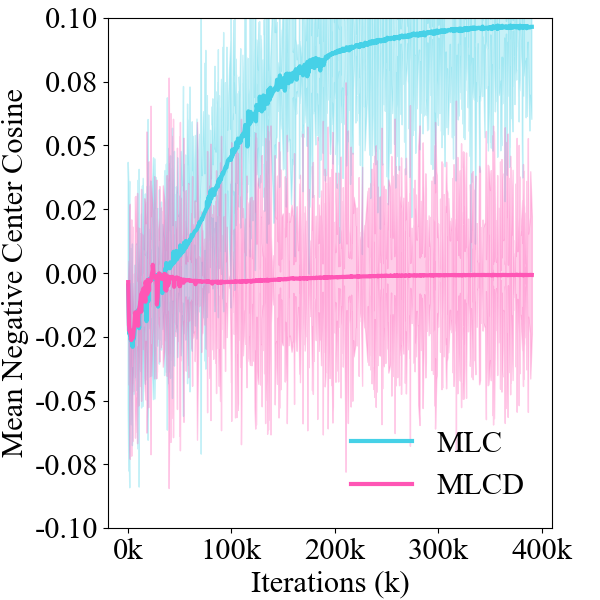}
    \caption{Mean $s_j$}
    \label{cosine:sj}
  \end{subfigure}
  \caption{Intra-class and inter-class similarity score comparisons between MLC and MLCD. Here, MLC and MLCD are trained on the LAION-400M dataset with the ViT-B/32 as the backbone and a batch size of 32K. (a) and (b) showcase histograms that compare the distributions of positive cosine similarities $\{s_i\}$ between MLC and MLCD, with MLCD clearly showing tighter sample alignment to positive class centers. (c) demonstrates that MLCD consistently achieves higher mean positive cosine values than MLC over iterations, indicating enhanced intra-class compactness. (d) demonstrates MLCD’s effectiveness in reducing mean negative cosine values compared to MLC, which indicates a more orthogonal relationship between samples and their negative class centers. This greater orthogonality facilitated by MLCD contributes to enhanced class separability. These figures highlight MLCD’s advanced capability in refining feature spaces for more distinct representation compared to MLC.}
  \label{fig:mlc_vs_mlcd}
  \vspace{-4mm}
\end{figure}

\noindent{\bf Multi-label Classification.}
As illustrated in Fig.~\ref{fig:pruityconflict}, a single image can encompass several visual components (\eg, objects or attributes). This implies that the single-class label may not cover all visual cues present in the image. 
To consider the different granularities of visual information
for each sample, we perform a similarity-based sorting against one million class centers, selecting the top $l$ class centers as the positive class centers for that sample. 
During training, this sample will be directed to move closer to these $l$ positive class centers, while simultaneously distancing from the other $k-l$ negative class centers. 
As shown in Fig.~\ref{fig1:multicluster}, our method assigns multiple class labels to each training example, capturing different granularities of visual signals in one image.

The corresponding similarity scores are represented as $\{s_i\}\, (i=1,2,\cdots,l)$ and $\{s_j\}\,(j=1,2,\cdots,k-l)$, respectively. 
To minimize each $s_j$ ($ \forall j \in \{1,2,\cdots,k-l\}$) as well as to maximize $s_i$ ($\forall i\in \{1,2,\cdots,l\}$), we employ a multi-label classification strategy~\cite{li2017improving,sun2020circle}. This is achieved by
\vspace{-2mm}
\begin{align}
\label{eqn:mlc}
\small
\mathcal{L}_\mathrm{MLC}  =  \log(1+ \underbrace{\sum_{j=1}^{k-l} \sum_{i=1}^l \exp(s_j-s_i))}_{contrastive}) 
 = \log(1+\underbrace{\sum_{j \in \Omega_{n}}\exp({s_j})\sum_{i \in \Omega_{p}}\exp({-s_i})}_{contrastive}), 
\end{align}
where $\Omega_{n}$ and $\Omega_{p}$ denote the negative and positive class set to simplify the representation. Eq.~\ref{eqn:mlc} iterates through every similarity pair to reduce $(s_j-s_i)$. To alleviate inter-class conflict as in~\cite{an2022killing,an2023unicom}, we also employ negative class sampling into Eq.~\ref{eqn:mlc}. Therefore, the loss is changed from $\log(1+\sum_{j \in \Omega_{n}}\exp({s_j})\sum_{i \in \Omega_{p}}\exp({-s_i}))$ to $\log(1+\sum_{j \in \Omega'_{n}}\exp({s_j})\sum_{i \in \Omega_{p}}\exp({-s_i}))$, where $\left |\Omega'_{n}  \right |= \left |\Omega_{n}  \right |*r$, and $r \in\left [0,1 \right ]$ is the negative class sampling ratio. $\Omega'_{n}$ is a subset of $\Omega_n$ that is randomly sampled during each loss calculation step.

\noindent{\bf Multi-label Classification Disambiguation.}
Optimizing $(s_j-s_i)$ usually leads to a decision boundary of $s_j-s_i=m$ ($m$ is the margin). 
However, this decision boundary allows ambiguity as indicated in Circle loss~\cite{sun2020circle}. 
For example, $\{s_j,s_i\}=\{0.1,0.4\}$ and $\{s'_j,s'_i\}=\{0.5,0.8\}$ both achieve the margin $m=0.3$. However, the gap between $s_i$ and $s'_j$ is only $0.1$, compromising the separability of the feature space. 

As we expect to maximize the within-class similarity $s_i$ and to minimize the between-class similarity $s_j$, we further introduce these two items into the multi-label classification loss:
\vspace{-2mm}
\begin{align}
\label{eqn:mlc2}
\mathcal{L}_\mathrm{MLCD}
& = \log(1+\underbrace{\sum_{j\in\Omega'_{n}}\exp({s_j})\sum_{i\in\Omega_{p}}\exp({-s_i})}_{contrastive}+\underbrace{\sum_{j\in\Omega'_{n}}\exp({s_j})}_{negative}+\underbrace{\sum_{i\in\Omega_{p}}\exp({-s_i})}_{positive}
) \notag \\ 
& = \log(1+\sum_{i \in \Omega_{p}}\exp({-s_i}))+ \log(1+\sum_{j \in \Omega'_{n}}\exp({s_j})), 
\end{align}
where $\Omega_p$ symbolizes the collection of positive class labels for each sample, $s_i$ encapsulates the score associated with each positive class, $\Omega'_n$ denotes the collection of negative class labels for each sample, and $s_j$ corresponds to the score for each negative class.
In Eq.~\ref{eqn:mlc2}, loss from positive class labels $\log(1+\sum_{i \in \Omega_{p}}\exp({-s_i}))$ and loss from negative class labels $\log(1+\sum_{j \in \Omega'_{n}}\exp({s_j}))$ are elegantly separated.
In Fig.~\ref{cosine:mlc} and Fig.~\ref{cosine:mlcd}, we compare the dynamic distributions of $s_i$ of MLC (Eq.~\ref{eqn:mlc}) and MLCD (Eq.~\ref{eqn:mlc2}) during training steps. Besides, Fig.~\ref{cosine:si} illustrates the average $s_i$ from MLC and MLCD during training. As we can see, the item designed for maximizing the within-class similarity $s_i$ in Eq.~\ref{eqn:mlc2} can significantly increase the intra-class cosine similarities, enhancing the intra-class compactness. In Fig.~\ref{cosine:sj}, the item designed for minimizing the between-class similarity $s_j$ can effectively suppress the inter-class cosine similarities, enforcing the inter-class discrepancy.

\section{Experiments}
\subsection{Experimental Setting}
Our models are pre-trained on the LAION-400M dataset~\cite{schuhmann2021laion} with the same model configurations as CLIP. The training process consists of 32 epochs, utilizing a batch size of 32K on 80 NVIDIA A100 GPUs. To expedite the training, we employ mixed-precision computation~\cite{micikevicius2017mixed} and flash attention~\cite{dao2023flashattention2}, while leveraging the DALI library for efficient data loading and pre-processing. We use the AdamW optimizer with a learning rate of $0.001$ and weight decay of $0.2$. To assess the performance of zero-shot classification and zero-shot image-text retrieval tasks, we employ contrastive learning to train a text encoder from scratch for 32 epochs with a frozen image encoder following Locked-image Tuning (LiT)~\cite{zhai2022lit}. The structure of the text encoder is also identical to CLIP. In the following experiments, unless otherwise specified, the model used is ViT-L/14, the number of classes ($k$) is one million, the ratio of sampled negative class centers ($r$) is $0.1$, and the number of positive labels ($l$) assigned to each image is $8$.

\subsection{Linear Probe}

Following the same evaluation setting as CLIP, we report the linear probe performance of our method on 26 datasets. As depicted in Tab.~\ref{tab:linear-probe-big-table}, inherent biases exist in different pre-training data. The WIT dataset is beneficial for action-related datasets (\eg, Kinetics700, UCF101), while LAION exhibits superior proficiency in object datasets (\eg, Cars, Birdsnap). Nevertheless, our method still achieves an average improvement of 1.1\% compared to CLIP. 
To isolate the confounding effects of pre-training data, we compare our model with OPENCLIP and UNICOM by using the LAION-400M dataset as the training data.
As shown in Fig.~\ref{fig:openclip_linear}, our method outperforms OPENCLIP on 25 datasets, demonstrating an average improvement of 2.3\%. 
In Fig.~\ref{fig:unicom_linear},
our model surpasses UNICOM on 23 datasets and achieves an average improvement of 1.3\%, confirming the effectiveness of the proposed multi-label loss.

\subsection{Zero-shot Classification}

\begin{figure}[t!]
\centering
\begin{subfigure}[b]{0.328\textwidth}
  \centering
  \includegraphics[width=\textwidth]{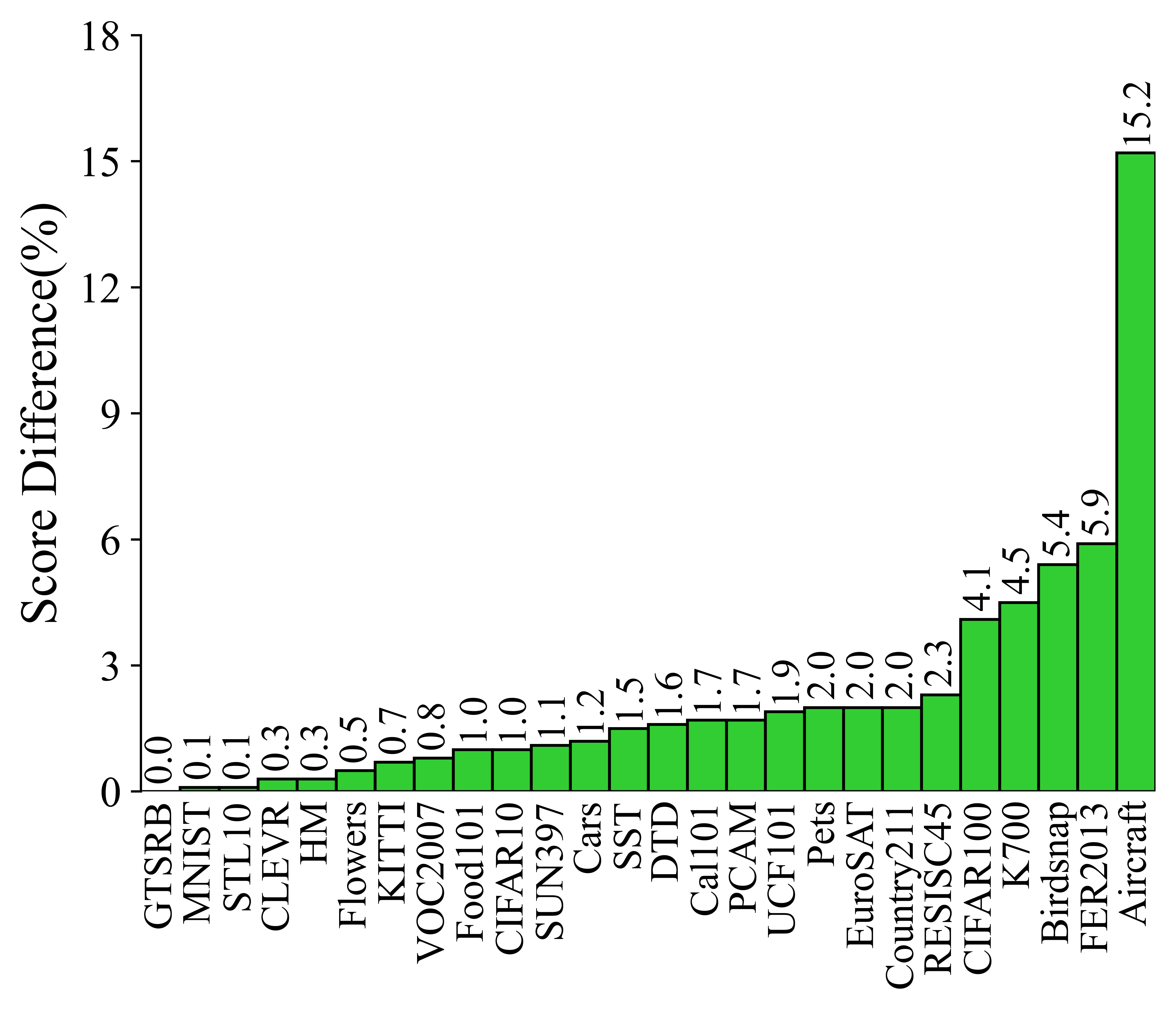}
  \captionsetup{font=tiny} 
  \caption{MLCD vs. OpenCLIP on linear}
  \label{fig:openclip_linear}
\end{subfigure}
\hfill
\begin{subfigure}[b]{0.328\textwidth}
  \centering
  \includegraphics[width=\textwidth]{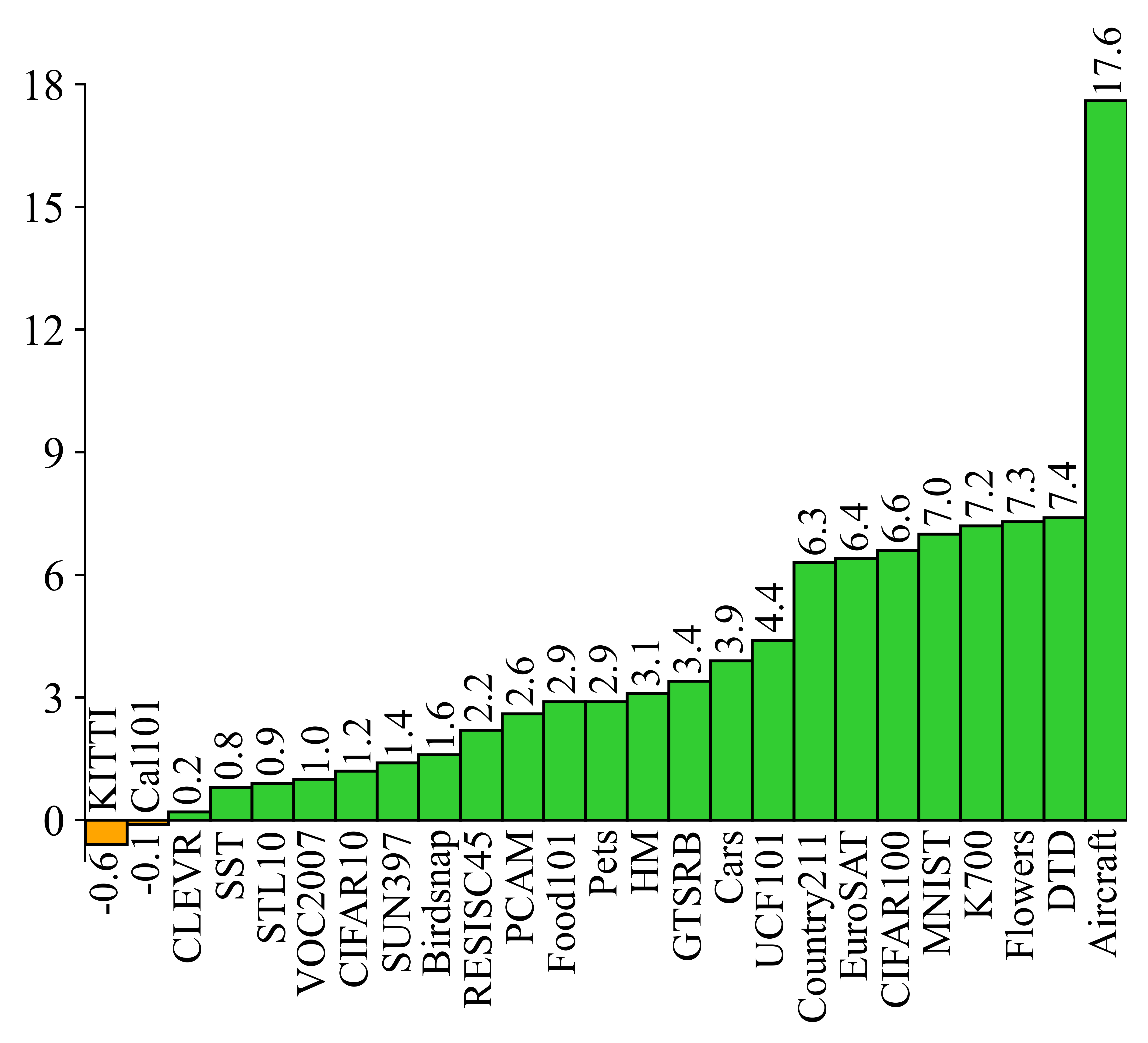}
  \captionsetup{font=tiny} 
  \caption{MLCD vs. OpenCLIP on zero-shot}
  \label{fig:openclip_zeroshot}
\end{subfigure}
\hfill
\begin{subfigure}[b]{0.328\textwidth}
  \centering
  \includegraphics[width=\textwidth]{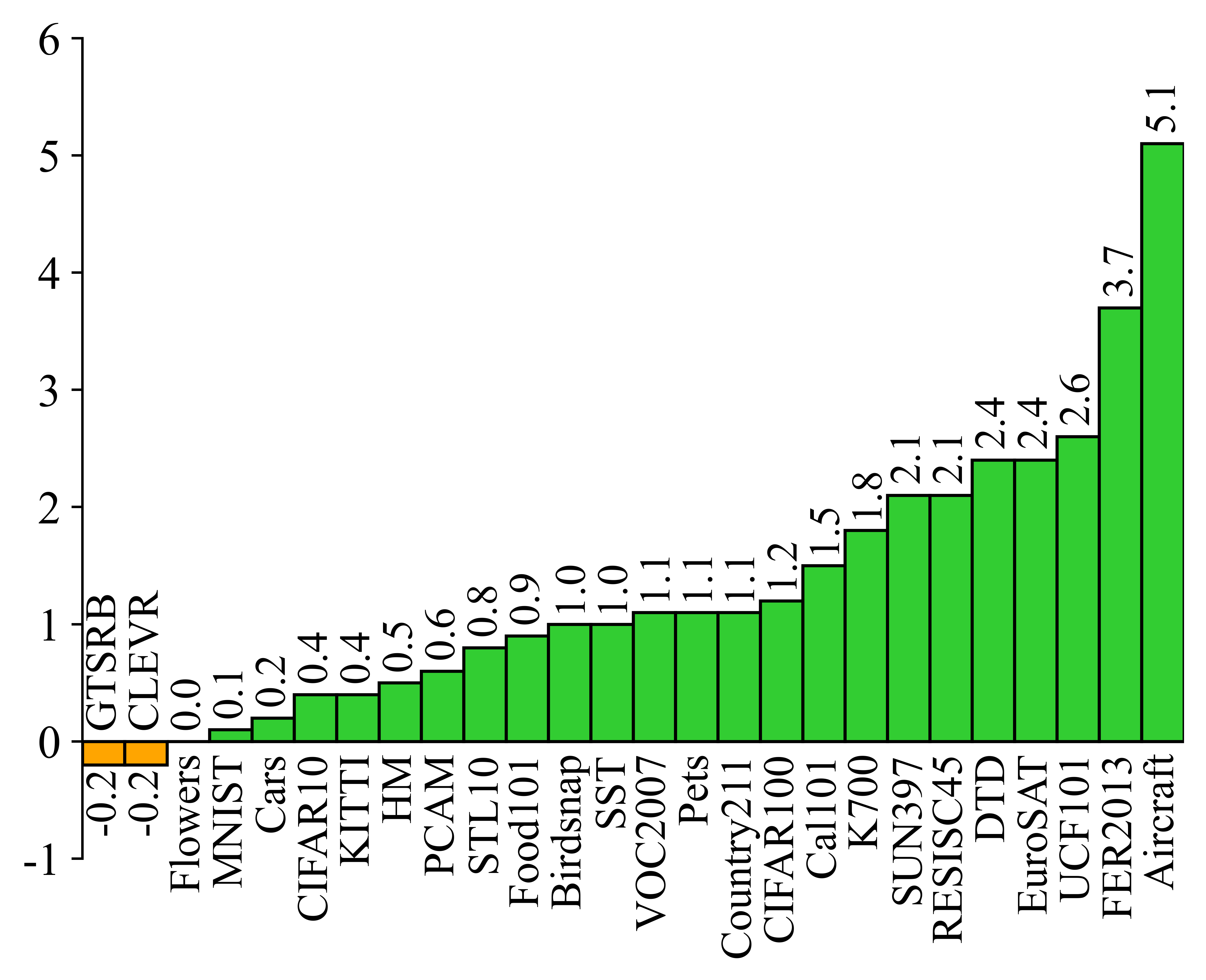}
  \captionsetup{font=tiny} 
  \caption{MLCD vs. UNICOM on linear}
  \label{fig:unicom_linear}
\end{subfigure}
\caption{Linear probe and zero-shot comparisons on different downstream datasets. The Y-axis shows the performance difference. Green bars indicate our model outperforms the baselines, while the orange bars depict our model is surpassed by the baselines.}
\label{fig:fig_histogram}
\vspace{-4mm}
\end{figure}

In Tab.~\ref{tab:zeroshot}, we present a comparison of our method with state-of-the-art approaches in zero-shot classification on 25 datasets. The prompt templates and class names are consistent with previous works~\cite{li2023scaling}. As depicted in Fig.~\ref{fig:openclip_zeroshot}, our method surpasses OpenCLIP on 23 datasets with 3.9\% average performance improvement. Although FLIP uses masking to save memory footprint to learn more samples per iteration, our method demonstrates better results on 15 out of 25 datasets in Tab.~\ref{tab:zeroshot} and achieves a significant performance boost of 1.5\% on average.

\begin{table}[t]
    \centering
    \caption{Linear probe performance of various pre-trained models on 26 datasets. $\dag$: Results reported in CLIP paper. $\ddag$: Results we reproduced. Entries in green are the best results using LAION-400M. Here, all methods employ the same backbone of ViT-L/14.}
    \vspace{-4mm}
    \label{tab:linear-probe-big-table}
    \resizebox{1.0\linewidth}{!}{
        \tablestyle{0.2mm}{1.1}
        \begin{tabular}{*l^l|^c^c^c^c^c^c^c^c^c^c^c^c^c^c^c^c^c^c^c^c^c^c^c^c^c^c^c}
            CASE & DATA & \datatag{Food101} & \datatag{{CIFAR10}} & \datatag{{CIFAR100}} & \datatag{{Birdsnap}} & \datatag{{SUN397}} & \datatag{{Cars}} & \datatag{{Aircraft}} & \datatag{{VOC2007}} & \datatag{{DTD}} & \datatag{{Pets}} & \datatag{{Cal101}} & \datatag{{Flowers}} & \datatag{{MNIST}} & \datatag{{FER2013}} & \datatag{{STL10}} & \datatag{{EuroSAT}} & \datatag{{RESISC45}} & \datatag{{GTSRB}} & \datatag{{KITTI}} & \datatag{{Country211}} & \datatag{{PCAM}} & \datatag{{UCF101}} & \datatag{{K700}} & \datatag{{CLEVR}} & \datatag{{HM}} & \datatag{{SST}} & \datatag{{AVG}}\\
            \midrule

            \rowstyle{\color{dt}} CLIP$^{\dag}$&WIT-400M& 95.2 &98.0 &87.5 &77.0 &81.8 &90.9 &69.4 &89.6 &82.1 &95.1 &96.5 &99.2 &99.2 &72.2 &99.8 &98.2 &94.1 &92.5 &64.7 &42.9 &85.8 &91.5 &72.0 &57.8 &76.2 &80.8 &84.2 \\
            \rowstyle{\color{dt}} CLIP$^{\ddag}$&WIT-400M&95.3 &98.1 &87.2 &77.8 &81.5 &90.7 &68.0 &89.7 &80.9 &94.9 &96.0 &99.2 &99.2 &72.3 &99.8 &96.7 &94.5 &92.9 &65.9 &41.9 &85.3 &91.0 &70.6 &59.6 &61.8 &79.8 &83.5 \\
            OPNCLIP$^{\ddag}$&LAION-400M&93.3 &97.9 &87.9 &78.0 &81.0 &93.6 &64.4 &91.7 &83.0 &93.3 &95.5 &98.8 &99.2 &66.5 &99.2 &97.1 &92.4 &92.5 &77.5 &32.5 &84.3 &88.1 &64.0 &59.8 &57.6 &71.9 &82.3 \\
            UNICOM&LAION-400M&93.4 &98.5 &90.8 &82.4 &80.0 &94.6 &74.5 &91.4 &82.2 &94.2 &95.7 &\better{99.3} &99.2 &68.7 &98.5 &96.7 &92.6 &\better{92.7} &77.8 &33.4 &85.4 &87.4 &66.7 &\better{60.3} &57.4 &72.4 &83.3 \\
            Ours&LAION-400M&\better{94.3} &\better{98.9} &\better{92.0} &\better{83.4} &\better{82.1} &\better{94.8} &\better{79.6} &\better{92.5} &\better{84.6} &\better{95.3} &\better{97.2} &\better{99.3} &\better{99.3} &\better{72.4} &\better{99.3} &\better{99.1} &\better{94.7} &92.5 &\better{78.2} &\better{34.5} &\better{86.0} &\better{90.0} &\better{68.5} &60.1 &\better{57.9} &\better{73.4} &\better{84.6} \\
            \bottomrule
        \end{tabular}
    }
\end{table}

\begin{table}[t!]
    \caption{Zero-shot classification performance on 25 datasets. $\dag$: Results reported in CLIP paper. $\ddag$: Results reported in FLIP paper. Entries in green are the best results using LAION-400M. Here, all methods employ the same backbone of ViT-L/14.}
    \label{tab:zeroshot}
    \vspace{-4mm} 
    \centering
    \resizebox{1.0\linewidth}{!}{
        \tablestyle{0.2mm}{1.1}
        \begin{tabular}{*l^l|^c^c^c^c^c^c^c^c^c^c^c^c^c^c^c^c^c^c^c^c^c^c^c^c^c^c^c}
            CASE & DATA & \datatag{Food101} & \datatag{{CIFAR10}} & \datatag{{CIFAR100}} & \datatag{{Birdsnap}} & \datatag{{SUN397}} & \datatag{{Cars}} & \datatag{{Aircraft}} & \datatag{{VOC2007}} & \datatag{{DTD}} & \datatag{{Pets}} & \datatag{{Cal101}} & \datatag{{Flowers}} & \datatag{{MNIST}} & \datatag{{STL10}} & \datatag{{EuroSAT}} & \datatag{{RESISC45}} & \datatag{{GTSRB}} & \datatag{{KITTI}} & \datatag{{Country211}} & \datatag{{PCAM}} & \datatag{{UCF101}} & \datatag{{K700}} & \datatag{{CLEVR}} & \datatag{{HM}} & \datatag{{SST}} & \datatag{{AVG}} \\
            \midrule

            \rowstyle{\color{dt}}CLIP$^{\dag}$&WIT-400M&92.9 &96.2 &77.9 &48.3 &67.7 &77.3 &36.1 &84.1 &55.3 &93.5 &92.6 &78.7 &87.2 &99.3 &59.9 &71.6 &50.3 &23.1 &32.7 &58.8 &76.2 &60.3 &24.3 &63.3 &64.0 &66.9 \\
            \rowstyle{\color{dt}}CLIP$^{\ddag}$&WIT-400M&91.0 &95.2 &75.6 &51.2 &66.6 &75.0 &32.3 &83.3 &55.0 &93.6 &92.4 &77.7 &76.0 &99.3 &62.0 &71.6 &51.6 &26.9 &30.9 &51.6 &76.1 &59.5 &22.2 &55.3 &67.3 &65.6 \\
            OpenCLIP$^{\ddag}$&LAION-400M&87.4&94.1 &77.1 &61.3 &70.7 &86.2 &21.8 &83.5 &54.9 &90.8 &\better{94.0} &72.1 &71.5 &98.2 &53.3 &67.7 &47.3 &29.3 &21.6 &51.1 &71.3 &50.5 &22.0 &55.3 &57.1 &63.6 \\
            FLIP$^{\ddag}$&LAION-400M&89.3&\better{97.2} &\better{84.1} &\better{63.0} &\better{73.1} &\better{90.7} &29.1 &83.1 &60.4 &92.6 &93.8 &75.0 &\better{80.3} &98.5&53.5 &\better{70.8} &41.4 &\better{34.8} &23.1 &50.3 &74.1 &55.8 &\better{22.7} &54.0 &\better{58.5} &66.0 \\
            Ours&LAION-400M&\better{90.3} &95.3 &83.7 &62.9 &72.1 &90.1 &\better{39.4} &\better{84.5} &\better{62.3} &\better{93.7} &93.9 &\better{79.4} &78.5 &\better{99.1} &\better{59.7} &69.9 &\better{50.7} &28.7 &\better{27.9} &\better{53.7} &\better{75.7} &\better{57.7} &22.2 &\better{58.4} &57.9 &\better{67.5} \\

            \bottomrule
        \end{tabular}} 
\end{table}

\subsection{Zero-shot Retrieval}

Tab.~\ref{tab:retrieval} reports zero-shot image-text retrieval results on Flickr30k and MSCOCO. In comparison to OpenCLIP, our model achieves 60.8\%/44.5\% I2T/T2I retrieval Recall@1 on the MSCOCO dataset, which is 2.8\%/3.2\% higher than OpenCLIP. Similarly, our model demonstrates significant improvements of 1.8\%/3.9\% on the Flickr30k dataset. Furthermore, compared to FLIP, our model exhibits either competitive or superior retrieval performance.

\begin{table}[t]
    \centering
    \caption{Zero-shot image-text retrieval on the test splits of Flickr30k and MSCOCO. $\ddag$: Results reported in FLIP paper. Entries in green are the best results using LAION-400M. Here, all methods employ the same backbone of ViT-L/14.}
    \label{tab:retrieval}
    \vspace{-4mm}
    \resizebox{0.85\linewidth}{!}{
        \tablestyle{3pt}{1.05}
        \begin{tabular}{*l^l^c^c^c^c^c^c^c^c^c^c^c^c}
            \toprule
             &  & \multicolumn{6}{c}{Text retrieval} & \multicolumn{6}{c}{Image retrieval} \\
             &  & \multicolumn{3}{c}{Flickr30k} & \multicolumn{3}{c}{MSCOCO} & \multicolumn{3}{c}{Flickr30k} & \multicolumn{3}{c}{MSCOCO} \\
            CASE & DATA & R@1 & R@5 & R@10 & R@1 & R@5 & R@10 & R@1 & R@5 & R@10 & R@1 & R@5 & R@10 \\ \midrule

            \rowstyle{\color{dt}} CLIP$^{\ddag}$ & WIT-400M & 87.8 & 99.1 & 99.8 & 56.2 & 79.8 & 86.4 & 69.3 & 90.2 & 94.0 & 35.8 & 60.7 & 70.7 \\
            OpenCLIP$^{\ddag}$ & LAION-400M & 87.3 & 97.9 & 99.1 & 58.0 & 80.6 & 88.1 & 72.0 & 90.8 & 95.0 & 41.3 & 66.6 & 76.1 \\
            FLIP$^{\ddag}$     & LAION-400M & \better{89.1} & \better{98.5} & \better{99.6} & 60.2 & 82.6 & 89.9 & 75.4 & 92.5 & 95.9 & 44.2 & 69.2 & 78.4 \\
            Ours               & LAION-400M & \better{89.1} & 98.4 & 99.5 & \better{60.8} & \better{83.2} & \better{91.3} & \better{75.9} & \better{93.1} & \better{96.8} & \better{44.5} & \better{69.6} & \better{79.9} \\

            \bottomrule
        \end{tabular}}

\end{table}
\setlength{\fboxsep}{1.5pt}

\begin{table}[t]
    \caption{ImageNet results under finetuning, linear probe, zero-shot, and zero-shot robustness evaluation settings. $\ddag$: Results reported in FLIP paper. Entries in green are the best results using LAION-400M. Here, all methods employ the same backbone of ViT-L/14.}    
    \label{tab:imagenet_robust}
    \vspace{-4mm}
    \centering
    \resizebox{1.0\linewidth}{!}{
        \tablestyle{4pt}{1.05}
        \begin{tabular}{*l^l|^c^c^c|^c^c^c^c^c }
        \toprule
            CASE & DATA & Finetune  & Linear Probe  &  Zero Shot & IN-V2 & IN-A & IN-R & ObjectNet & IN-Sketch \\
            \midrule
            \rowstyle{\color{dt}}CLIP$^{\ddag}$ & WIT-400M & -       & 83.9        & 75.3 & 69.5 & 71.9 & 86.8 & 68.6 & 58.5 \\
            OpenCLIP$^{\ddag}$ & LAION-400M & 86.2 &  82.1 &   72.8  & 64.0 & 48.3 & 84.3 & 58.8 & 56.9 \\
            FLIP$^{\ddag}$ & LAION-400M & -      & -       & 74.6 & 66.8 & 51.2 & \better{86.5} & 59.1 & 59.9 \\
            \midrule
            Ours & LAION-400M & \better{87.1}  & \better{84.6}     &\better{75.6} & \better{68.9} & \better{56.4} & 85.1 & \better{62.7} & \better{60.4} \\
        \bottomrule
        \end{tabular}}
\end{table}

\subsection{ImageNet Classification and Robustness Evaluation}

We evaluate performance on ImageNet~\cite{deng2009imagenet} under three distinct settings: finetuning, linear probe, and zero-shot. As shown in Tab.~\ref{tab:imagenet_robust}, our ViT-L/14 model achieves better performance on all settings, outperforming OpenCLIP by $0.9\%$ under the finetuning setting, and surpassing FLIP by $1.0\%$ under the zero-shot setting. These improvements indicate that multi-label cluster discrimination can better encode the semantics of data than instance discrimination. Following FLIP~\cite{li2023scaling}, we conduct a robustness evaluation as shown in Tab.~\ref{tab:imagenet_robust}. In comparison to the models pre-trained on LAION, our method demonstrates superior robustness compared to both OpenCLIP and FLIP. It is worth noting that the performance gap between our model pre-trained on LAION and CLIP pre-trained on WIT arises from the statistical differences in pre-training data.

\subsection{Ablation Study}

\begin{table}[t]
    \caption{\textbf{Ablation experiments}. The model backbone used here is ViT-B/32. Pre-training is executed on the LAION-400M dataset for a duration of 5 epochs. Performance assessment is undertaken using a linear probe on the ImageNet validation set.}
    \vspace{-2mm}
    \label{tab:ablationpara}
    \tablestyle{3pt}{1.05}
    \begin{subtable}{.495\textwidth}
        \resizebox{\linewidth}{!}{
            \begin{tabular}{y{80}|x{25}x{25}x{25}x{25}x{25}x{25}}
            \toprule
                 Num Classes & 100K  & 200K  & 500K  & 1M            & 2M   & 5M   \\
                \midrule
                IN1K       & 66.9 & 71.1 & 74.4 & \better{75.2} & 74.9 & 74.7 \\
            \bottomrule
            \end{tabular}}
        \caption{The number of \textbf{classes} in training set.}
        \label{tab:abl1}
    \end{subtable}%
    \hfill
    \begin{subtable}{.495\textwidth}
        \resizebox{\linewidth}{!}{
            \begin{tabular}{y{80}|x{25}x{25}x{25}x{25}x{25}x{25}}
            \toprule
                Sampling Ratio & 0.01 & 0.05 & 0.1           & 0.2  & 0.5  & 1.0  \\
                \midrule
                IN1K         & 73.4 & 75.1 & \better{75.2} & 74.9 & 68.3 & 63.2 \\
            \bottomrule
            \end{tabular}}
        \caption{The \textbf{ratio} of negative class centers.}
        \label{tab:abl2}
    \end{subtable}%
    \vspace{-0.1cm}
    \begin{subtable}{.495\textwidth}
        \resizebox{\linewidth}{!}{
            \begin{tabular}{y{80}|x{25}x{25}x{25}x{25}x{25}x{25}}
            \toprule
                Positive Centers & 1    & 2    & 4    & 8             & 16   & 32   \\
                \midrule
                IN1K             & 71.4 & 72.9 & 73.2 & \better{75.2} & 72.1 & 68.7 \\
                \bottomrule
            \end{tabular}
        }
        \caption{The effect of \textbf{multi labels} per sample.}
        \label{tab:abl3}
    \end{subtable}%
    \hfill
    \begin{subtable}{.495\textwidth}
        \resizebox{\linewidth}{!}{
            \begin{tabular}{y{80}|x{25}x{25}x{25}x{25}x{25}x{25}}
                \toprule
                Positive Threshold & 0.95 & 0.93 & 0.91          & 0.89 & 0.87 & 0.85 \\
                \midrule
                IN1K               & 72.2 & 72.7 & \better{73.3} & 72.4 & 68.7 & 63.2 \\
                \bottomrule
            \end{tabular}
        }
        \caption{The effect of \textbf{positive thresholds}.}
        \label{tab:abl4}
    \end{subtable}%

\end{table}

\noindent{\bf Number of Classes.}
The number of classes ($k$) plays a crucial role in balancing inter-class conflict and intra-class purity. In Tab.~\ref{tab:abl1}, we observe that as the number of classes increases from 100K to 1M, there is a gradual increase in intra-class purity, leading to an improved performance on ImageNet. However, as the number of classes continues to increase from 1M to 5M, inter-class conflicts gradually escalate, resulting in a deteriorated performance. 

\noindent{\bf Inter-class sampling Ratio.}
The inter-class sampling ratio ($r$) influences the number of negative samples and directly affects the likelihood of encountering inter-class conflicts. A sample ratio of $0.01$ yields a linear probe performance of only $73.4\%$ due to the limited number of negative samples, which adversely affects the representation learning. Conversely, a sample ratio of $1.0$ substantially increases the probability of encountering inter-class conflicts. Tab.~\ref{tab:abl2} presents that the superior linear probe performance of $75.2\%$ is achieved when employing a sample ratio of $0.1$.

\noindent{\bf Multi-label Assignment.}
We explore two different approaches to obtain multi-labels. Firstly, we artificially assign a predetermined number of labels to each sample. Tab.~\ref{tab:abl3} presents linear probe results on ImageNet with different numbers of positive centers. Consequently, we observe a gradual improvement in performance as the number of positive centers increases from 1 to 8. However, as the number of positive centers continues to increase, the inclusion of excessive positive centers introduces noise labels, leading to a degradation in performance.
Additionally, we have also investigated the use of sample-cluster similarity thresholds to obtain multiple labels. This approach results in varying numbers of positive centers associated with each sample. However, as shown in Tab.~\ref{tab:abl4}, the performance of applying adaptive positive centers is generally lower compared to that of using fixed assignment of positive centers (Tab.~\ref{tab:abl3}). This indicates that the global similarity threshold is hard to search while the fixed assignment strategy benefits from the prior that the daily image statistically contains several visual concepts.
 
\noindent{\bf Effectiveness of MLCD Compared to MLC.}
In Tab.~\ref{tab:mlcvsmlcd}, we compare the performance of the vanilla MLC (Eq.~\ref{eqn:mlc}) and the proposed MLCD (Eq.~\ref{eqn:mlc2}) on the ImageNet. Both MLC and MLCD employ the negative class center sampling with a ratio of $0.1$. MLCD outperforms MLC in all three settings, confirming the effectiveness of the two additional optimization targets. 

\begin{table}[t!]
    \caption{Ablation experiments of the proposed contrastive loss decomposition. Pre-training is executed on the LAION-400M dataset by 32 epochs. The model backbone used here is ViT-B/32. Results are reported on the ImageNet validation dataset.}
    \label{tab:mlcvsmlcd}
    \vspace{-2mm}
    \centering
    \resizebox{0.60\linewidth}{!}{
        \tablestyle{4pt}{1.05}
        \begin{tabular}{y{24}y{60}|ccc}  
        \toprule
            CASE & DATA &  Finetune & Linear Probe & Zero Shot \\
            \midrule
            MLC  &  LAION-400M & 80.9 & 76.9 & 63.9 \\
            MLCD &  LAION-400M & \better{81.2} & \better{78.1} & \better{64.5}  \\  
            \bottomrule
        \end{tabular}}

\end{table}

\begin{figure}[!t] 
\centering
  \begin{subfigure}[b]{0.35\textwidth} 
    \centering 
    \includegraphics[width=\textwidth]{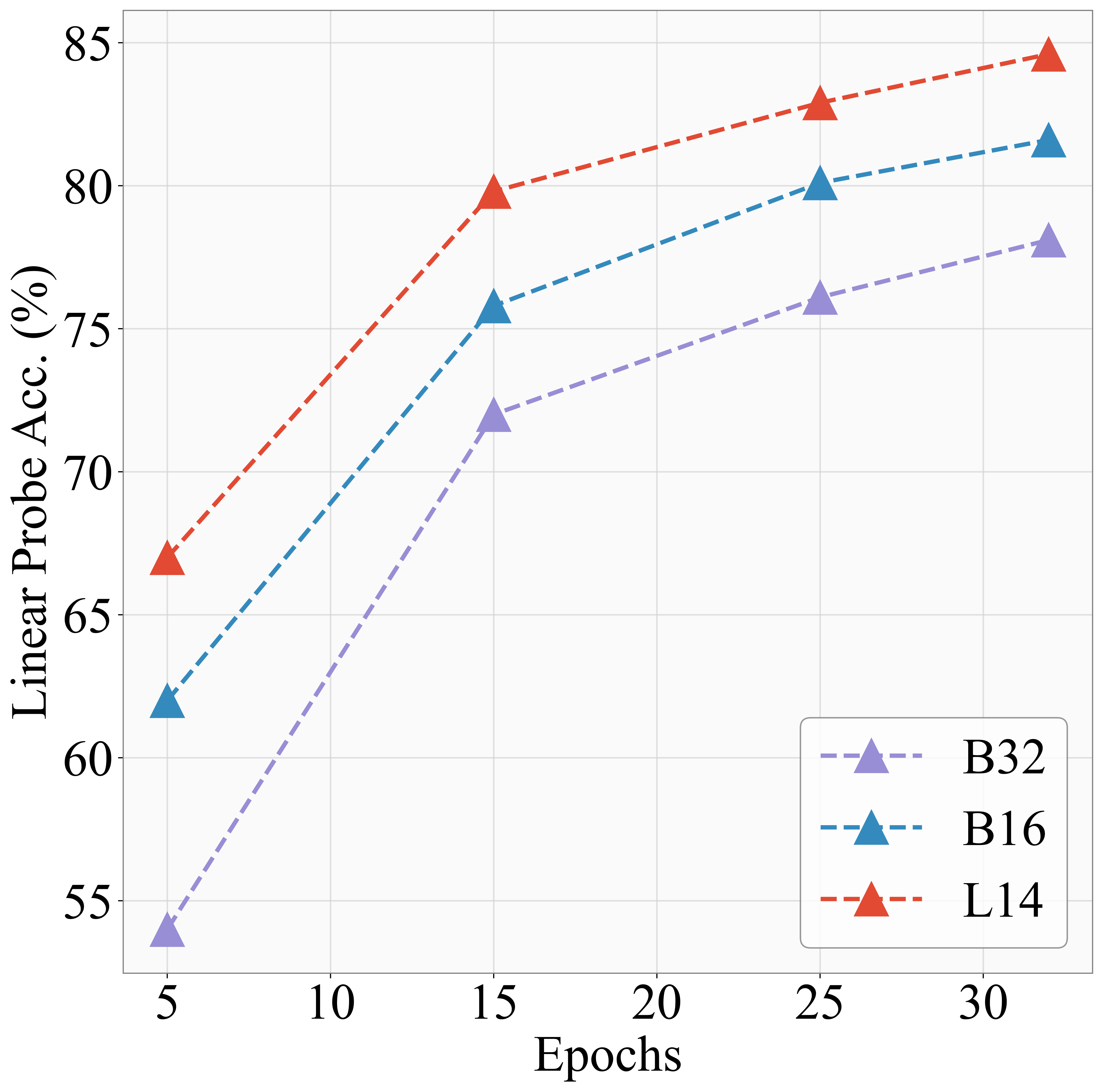}
    \caption{Performance vs. Epochs}
    \label{scale:0}
  \end{subfigure}
  \hspace{1cm} 
  \begin{subfigure}[b]{0.35\textwidth} 
    \centering 
    \includegraphics[width=\textwidth]{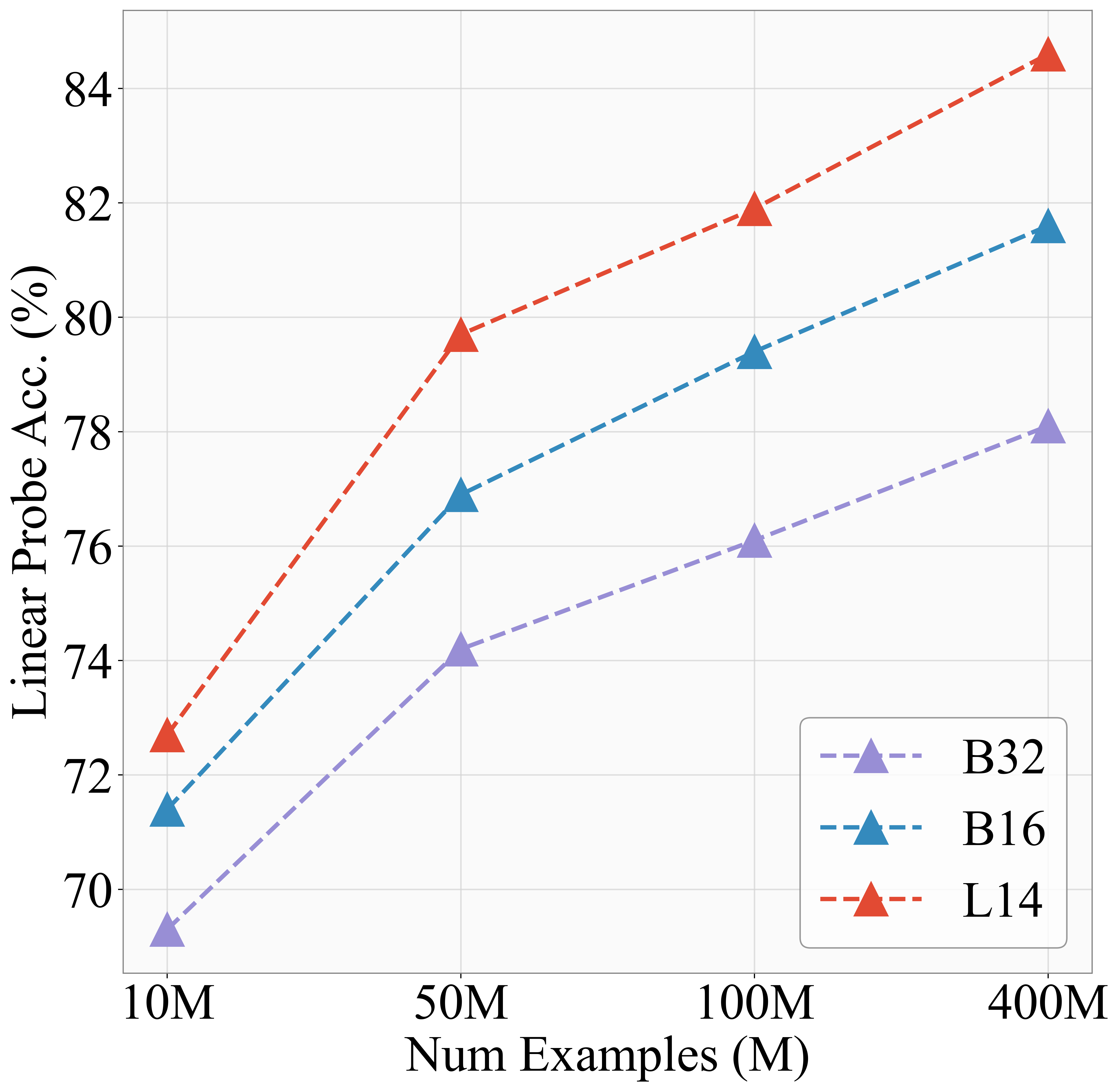}
    \caption{Performance vs. Num examples}
    \label{scale:1}
  \end{subfigure}
  \caption{(a) The convergence curves of different ViTs. (b) The scalability curves of different ViTs under varying dataset scales. Larger ViTs and datasets lead to better model performance.}
  \label{scale}
\end{figure}

\noindent{\bf Scalability.}
In Fig.~\ref{scale:0} and Fig.~\ref{scale:1}, we validate the scalability of our method. Scaling up the ViT model and incorporating more data can significantly enhance our model's performance.

\noindent{\bf Effectiveness of MLCD on Different Training Data.}
In Tab.~\ref{tab:laioncoyo}, we compare the linear probe performance of the proposed multi-label cluster discrimination approach (\ie, MLCD) and the single-label cluster discrimination method (\eg, UNICOM) on LAION-400M and COYO-700M. The hyper-parameter settings on COYO-700M follow the best settings on LAION-400M as explored in Tab.~\ref{tab:ablationpara}. 
As we can see from the results, the proposed MLCD consistently outperforms UNICOM by $2.2\%$ and $1.6\%$ when using LAION-400M and COYO-700M as the training data. In addition, the COYO-700M supports superior performance on action-related evaluation, achieving $3.3\%$ improvement on Kinetics700 by using MLCD.

\begin{table}[t]
    \caption{Comparisons of linear probe performance across 26 different datasets for models trained on LAION-400M and COYO-700M datasets. Here, all methods employ the same backbone of ViT-B/32.}
    \vspace{-4mm}
    \label{tab:laioncoyo}
    \centering
    \resizebox{1.0\linewidth}{!}{
        \tablestyle{0.2mm}{1.1}
        \begin{tabular}{*l^l|^c^c^c^c^c^c^c^c^c^c^c^c^c^c^c^c^c^c^c^c^c^c^c^c^c^c^c}
            CASE & DATA & \datatag{Food101} & \datatag{{CIFAR10}} & \datatag{{CIFAR100}} & \datatag{{Birdsnap}} & \datatag{{SUN397}} & \datatag{{Cars}} & \datatag{{Aircraft}} & \datatag{{VOC2007}} & \datatag{{DTD}} & \datatag{{Pets}} & \datatag{{Cal101}} & \datatag{{Flowers}} & \datatag{{MNIST}} & \datatag{{FER2013}} & \datatag{{STL10}} & \datatag{{EuroSAT}} & \datatag{{RESISC45}} & \datatag{{GTSRB}} & \datatag{{KITTI}} & \datatag{{Country211}} & \datatag{{PCAM}} & \datatag{{UCF101}} & \datatag{{K700}} & \datatag{{CLEVR}} & \datatag{{HM}} & \datatag{{SST}} & \datatag{{AVG}}\\
            \midrule

            UNICOM&LAION-400M&85.8 &96.8 &86.6 &70.2 &74.6 &93.3 &70.7 &88.3 &78.0 &93.1 &94.6 &98.5 &98.7 &64.3 &97.8 &96.8 &90.6 &90.0 &76.4 &22.5 &82.9 &84.2 &57.2 &52.6 &52.4 &62.1 &79.2 \\
            MLCD&LAION-400M&\better{87.8} &\better{97.5} &\better{88.2} &\better{72.4} &\better{77.6} &\better{93.8} &\better{71.4} &\better{91.9} &\better{80.4} &\better{93.2} &\better{96.9} &\better{98.8} &\better{99.3} &\better{66.4} &\better{98.6} &\better{98.6} &\better{92.1} &\better{90.5} &\better{77.7} &\better{30.9} &\better{83.4} &\better{86.3} &\better{60.9} &\better{54.1} &\better{57.9} &\better{70.4} &\better{81.4} \\
             \midrule
            UNICOM&COYO-700M &88.1 &95.4 &85.8 &71.4 &76.6 &93.1 &72.7 &88.1 &81.7 &93.3 &95.6 &97.5 &\better{99.3} &70.3 &98.7 &97.8 &91.5 &89.9 &76.7 &30.4 &82.1 &86.3 &61.8 &57.4 &64.3 &69.1 &81.3 \\
            MLCD&COYO-700M &\better{90.2} &\better{96.9} &\better{86.8} &\better{72.1} &\better{77.4} &\better{93.5} &\better{74.7} &\better{90.4} &\better{83.5} &\better{93.6} &\better{97.7} &\better{98.8} &\better{99.3} &\better{70.9} &\better{99.1} &\better{99.0} &\better{92.7} &\better{90.1} &\better{77.5} &\better{33.7} &\better{84.4} &\better{87.5} &\better{64.2} &\better{59.2} &\better{68.4} &\better{73.4} &\better{82.9} \\

            \bottomrule
        \end{tabular}
    }

\end{table}

\newcolumntype{S}{@{}>{\lrbox0}l<{\endlrbox}}  %
\definecolor{lightgreen}{HTML}{D8ECD1}

\noindent{\bf Effectiveness of MLCD in Vision Language Model.}
Tab.~\ref{tab:VLM} compares the performance of replacing the vision tower in LLaVA-1.5~\cite{liu2024llava_V1_5} from the CLIP model with our MLCD model. We validate the effectiveness of our MLCD under both Qwen2-7B and Qwen2-72B~\cite{qwen1,qwen2} settings across 14 test datasets. To align the experimental settings as in LLaVA-1.5, our model is fine-tuned for one epoch at a resolution of $336\times336$
after training at a resolution of $224\times224$.
It can be observed that our method, MLCD, outperforms CLIP on most of the test datasets. However, there is a noticeable drop in performance on OCR-related benchmarks, such as TextVQA~\cite{vlm_text_vqa} and AI2D~\cite{vlm_ai2d}, under both 7B and 72B settings. 
To this end, we will incorporate additional OCR models for clustering to enhance our OCR capabilities in the future.

\setlength{\fboxsep}{3pt}
\begin{table}[t]
\caption{Evaluation of different visual towers (\ie, CLIP and MLCD) used in VLM. The evaluation settings and test datasets align with LLaVA-1.5. The MLCD model (ViT-L/14) used here has employed training data from both LAION-400M and COYO-700M.}
\vspace{-2mm}
\label{tab:VLM}
\centering
\resizebox{1.0\linewidth}{!}{
\tablestyle{1.6pt}{1.2}
\begin{tabular}{*l^c|^c^c^c^c^c^c^c^c^c^c^c^c^c^c^c^c^c^c^c^c}
\toprule
\multirow{2}{*}{LLM}
 & \multirow{2}{*}{Vision Tower} & \multirow{1}{*}{VQAv2} & \multirow{1}{*}{GQA} & \multirow{1}{*}{VisWiz} & \multirow{1}{*}{SQA} & \multirow{1}{*}{TVQA}  & \multirow{1}{*}{L-Wild} & \multirow{1}{*}{AI2D} & \multirow{1}{*}{MathV} & \multirow{1}{*}{HBI} & 
\multirow{1}{*}{MMMU} & \multirow{1}{*}{cMMMU}  & \multicolumn{2}{c}{MMBench} & \multicolumn{3}{c}{SEED-Bench} & \multicolumn{2}{c}{MME} \\
\cline{3-20}
& & Val & Eval & Val & Img & Val & Test & Test & Mini  & ALL & Val & Val & EN & CN & All & Img & Vid & Per & Cog\\
\midrule
Qwen2-7B&CLIP&77.99 &62.66 &\better{48.58}&72.24 &\better{48.98}&58.70 &\better{64.86}&\better{33.60}&\better{39.96}&40.70 &\better{33.70}&72.03 &70.29 &64.25 &69.40 &44.72 &1512 &335 &\\
Qwen2-7B&Ours&\better{78.32}&\better{63.56}&46.27 &\better{74.22}&42.52 &\better{58.90}&62.82 &\better{33.60}&39.46 &\better{42.30}&33.10 &\better{73.88}&\better{71.47}&\better{65.79}&\better{71.05}&\better{45.89}&\better{1558}&\better{384}&\\
Qwen2-72B&CLIP&79.47 &63.81 &67.14 &\better{76.10}&\better{62.31}&65.41 &\better{72.41}&38.30 &45.10 &39.70 &37.45 &76.63 &75.39 &66.54 &72.28 &44.71 &1596 &378 &\\
Qwen2-72B&Ours&\better{79.51}&\better{66.80}&\better{67.37}&74.69 &57.32 &\better{66.00}&71.41 &\better{46.5}&\better{45.21}&\better{44.70}&\better{41.20}&\better{78.59}&\better{77.24}&\better{68.67}&\better{76.53}&\better{45.91}&\better{1633}&\better{383}&\\

\bottomrule
\end{tabular}
}
\end{table}

\begin{figure}
\centering
  \includegraphics[width=1.0\textwidth]{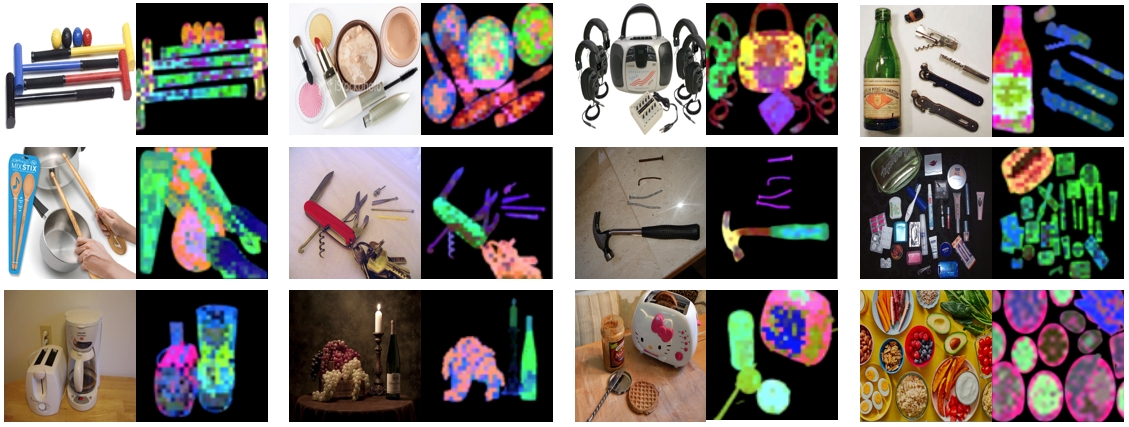}
\caption{PCA visualization of patch features extracted by our MLCD model. We fine-tuned the ViT-L/14 model on LAION-400M for one epoch at the resolution of $448\times448$, which allows each image to have $32\times32$ tokens for visualization. 
For each image, PCA is conducted on the extracted patch features to three principal components, which are subsequently normalized to the range of [0, 255] and mapped into the RGB space. Patches displaying similar colors indicate semantic similarities, reflecting that they embody analogous elements or attributes.}
\label{fig:pca}
\end{figure}

\noindent{\bf Semantic Visualization.}  In Fig.~\ref{fig:pca}, we show the results of the Principal Component Analysis (PCA) performed on the patch features extracted by our MLCD model. We fine-tune our ViT-L/14 model on the LAION-400M dataset by one epoch using the resolution of $448\times448$. As the patch size is $14\times14$, we can obtain $32\times32\times1024$ spatial-wise tokens for each image. Then, we build a PCA projection from $32\times32\times1024$ to $32\times32\times3$. After we threshold
the first component, we only keep patches with a positive value. 
As we can see from Fig.~\ref{fig:pca}, the unsupervised foreground/background detector, based on detecting the
highest variance direction, can separate the salient objects from the background. Afterward, we map the three PCA projection parameters into three different colors (\ie, [R, G, B]). 
As shown in Fig.~\ref{fig:pca}, objects from the same category exhibit color consistency, and objects from different categories present distinguishable colors, which indicates that the proposed multi-label cluster discrimination method can effectively capture multiple semantic signals from one image.

\section{Conclusions}
In this paper, we propose a novel multi-label cluster discrimination method to cope with multiple visual signals existing in one image. Compared to the vanilla version of the multi-label loss, which seeks to narrow the relative gap between inter-class similarities and intra-class similarities, our method introduces another two
optimization targets (\ie, decreasing inter-class similarities and increasing intra-class similarities) into the loss function. Introducing these two items enables the elegant separation of losses from positive and negative classes and alleviates the ambiguity on the decision boundary.
Extensive experimental results show that the proposed multi-label cluster discrimination loss is effective for providing better transferrable features on multiple downstream tasks than both instance and cluster discrimination methods.

\bibliographystyle{splncs04}
\bibliography{eccv2024}

\newpage
\appendix
\section{Appendix}

\subsection{Pre-training Details}

\subsubsection{Encoders}
Tab.~\ref{tab:vit} shows the architectures used in this paper. The designs follow CLIP~\cite{radford2021learning}. Our image encoders involve \mbox{ViT-B} and \mbox{ViT-L}, using the same patch size as in CLIP.

\subsubsection{Hyper-parameters}
Our default pre-training configuration is shown in Tab.~\ref{hyperparameter}. During the training process of the text encoder, the hyper-parameters are the same as those of the pre-training for the image encoder. The vision model is frozen, preventing any backpropagation of gradients. When calculating the multi-label contrastive loss, we follow the approaches of ArcFace~\cite{deng2019arcface} and Unicom~\cite{an2023unicom}, we apply L2 normalization to both the features and the class centers, and introduce a margin ($m=0.3$) for the positive classes.

\subsubsection{Downstream Datasets}  
We use 27 image classification datasets to prove the effectiveness of our method. These datasets
include Food101~\cite{bossard2014food}, CIFAR10~\cite{krizhevsky2009learning}, CIFAR100~\cite{krizhevsky2009learning}, 
Birdsnap~\cite{berg2014birdsnap},
SUN397~\cite{xiao2010sun},
Stanford Cars~\cite{KrauseStarkDengFei-Fei_3DRR2013},
FGVC Aircraft~\cite{maji2013fine},
VOC2007~\cite{everingham2007pascal},
DTD~\cite{cimpoi2014describing},
Pets~\cite{parkhi2012cats}, 
Caltech101~\cite{fei2004learning},
Flowers102~\cite{nilsback2008automated},
MNIST~\cite{lecun1998gradient},
FER2013~\cite{Goodfellow2013ChallengesIR},
SLT10~\cite{coates2011analysis},
EuroSAT~\cite{helber2019eurosat},
RESISC45~\cite{cheng2017remote},
GTSRB~\cite{stallkamp2012man},
KITTI~\cite{geiger2012we},
Country211~\cite{radford2021learning},
PCAM~\cite{veeling2018rotation},
UCF101~\cite{soomro2012ucf101},
Kinetics700~\cite{carreira2019short},
CLEVR~\cite{johnson2017clevr},
Hateful Memes~\cite{kiela2020hateful},
SST2~\cite{radford2021learning}, and
ImageNet~\cite{deng2009imagenet}. Details on each dataset and the corresponding evaluation metrics are provided in Tab.~\ref{linearprobedatasets}.

\begin{table}[h]
\vspace{2mm}
\caption{ViT hyper-parameters.}
\vspace{-4mm}
\label{tab:vit}
\centering
\tablestyle{2mm}{1.05}
\resizebox{1.0\linewidth}{!}{
\begin{tabular}{l|ccccccccc} \toprule
& Learning & Embedding & Input      & \multicolumn{3}{c}{Vision Transformer} & \multicolumn{3}{c}{Text Transformer} \\
Model & rate & dimension & resolution & layers & width & heads  & layers & width & heads \\ \midrule
ViT-B/32 & $5 \times 10^{-4}$ & 512 & 224 & 12 & 768 & 12 & 12 & 512 & 8 \\
ViT-B/16 & $5 \times 10^{-4}$ & 512 & 224 & 12 & 768 & 12 & 12 & 512 & 8 \\
ViT-L/14 & $4 \times 10^{-4}$ & 768 & 224 & 24 & 1024 & 16 & 12 & 768 & 12 \\
\bottomrule
\end{tabular}
}
\end{table}

\begin{table}[t]
\caption{Training hyper-parameters.}
\label{hyperparameter}
\centering
\tablestyle{5mm}{1.05}
\begin{tabular}{l|y{30}} \toprule
Hyperparameter  & Value \\ \midrule
Batch size & 32800 \\
Vocabulary size & 49408 \\
Training epochs & 32 \\
Maximum temperature & 100.0 \\
Weight decay & 0.2 \\
Warm-up iterations & 2000 \\
\bottomrule
\end{tabular}
\end{table}
\begin{table}[t!]
        \caption{List of linear probe datasets with the data distribution and evaluation metrics.}
    \label{linearprobedatasets}
    \centering
    \resizebox{0.99\linewidth}{!}{
        \tablestyle{5mm}{1.1}
        \begin{tabular}{llllr}
            \toprule
            \multicolumn{1}{l}{Dataset} & \multicolumn{1}{c}{Num Classes} & \multicolumn{1}{c}{Train size} & \multicolumn{1}{c}{Test size} & \multicolumn{1}{c}{Evaluation metric}\\
            \midrule
            LAION400M~\cite{schuhmann2021laion} & 1000000 & 389737314 & - & -\\
            COYO700M~\cite{kakaobrain2022coyo-700m}  & 1000000 & 686591232 & - & -\\
            \midrule
            Food101~\cite{bossard2014food} & 102 & 75,750 & 25,250 & accuracy\\
            CIFAR10~\cite{krizhevsky2009learning} & 10 & 50,000 & 10,000 & accuracy\\
            CIFAR100~\cite{krizhevsky2009learning} & 100 & 50,000 & 10,000 & accuracy\\
            Birdsnap~\cite{berg2014birdsnap} & 500 & 42,138 & 2,149 & accuracy\\
            SUN397~\cite{xiao2010sun} & 397 & 19,850 & 19,850 & accuracy\\
            Cars~\cite{KrauseStarkDengFei-Fei_3DRR2013} & 196 & 8,144 & 8,041 & accuracy\\
            Aircraft~\cite{maji2013fine} & 100 & 6,667 & 3,333 & mean per class\\
            VOC2007~\cite{everingham2007pascal} & 20 & 5011 & 4952 & 11-point mAP\\
            DTD~\cite{cimpoi2014describing} & 47 & 3,760 & 1,880 & accuracy\\
            Pets~\cite{parkhi2012cats} & 37 & 3,680 & 3,669 & mean per class\\
            Caltech101~\cite{fei2004learning} & 101 & 3,000 & 5,677 & mean-per-class\\
            Flowers~\cite{nilsback2008automated} & 102 & 2,040 & 6,149 & mean per class\\
            MNIST~\cite{lecun1998gradient} & 10 & 60,000 & 10,000 & accuracy\\
            FER2013~\cite{KrauseStarkDengFei-Fei_3DRR2013} & 8 & 32,140 & 3,574 & accuracy\\
            STL10~\cite{coates2011analysis} & 10 & 5,000 & 8,000 & accuracy\\
            EuroSAT~\cite{helber2019eurosat} & 10 & 10,000 & 5,000 & accuracy\\
            RESISC45~\cite{cheng2017remote} & 45 & 3,150 & 25,200 & accuracy\\
            GTSRB~\cite{stallkamp2012man} & 43 & 26,640 & 12,630 & accuracy\\
            KITTI~\cite{geiger2012we} & 4 & 6770 & 711 & accuracy\\
            Country211~\cite{radford2021learning} & 211 & 42,200 & 21,100 & accuracy\\
            PCAM~\cite{veeling2018rotation} & 2 & 294,912 & 32,768 & accuracy\\
            UCF101~\cite{soomro2012ucf101} & 101 & 9,537 & 1,794 & accuracy\\
            Kinetics700~\cite{carreira2019short} & 700 & 530,779 & 33,944 & mean(top1,top5)\\
            CLEVR~\cite{johnson2017clevr} & 8 & 2,000 & 500 & accuracy\\
            Memes~\cite{kiela2020hateful} & 2 & 8,500 & 500 & ROC AUC\\
            SST2~\cite{radford2021learning} & 2 & 7,792 & 1,821 & accuracy\\
            ImageNet~\cite{deng2009imagenet} & 1000 & 1,281,167 & 50,000 & accuracy\\
            \bottomrule
        \end{tabular}}
\end{table}

\begin{table}[t!]
\caption{Comparisons of linear probe performance across 27 different downstream datasets. Different models (\ie, UNICOM and MLCD) are trained on the automatically clustered ImageNet dataset with different class numbers. UNICOM employs a single label, while the proposed MLCD employs eight labels for each training sample. All methods use the same ResNet50~\cite{he2016deep,radford2021learning} architecture as the backbone.}
\label{tab:unicom_mlcd}
\centering
\resizebox{1.0\linewidth}{!}{
\tablestyle{0.6mm}{1.1}
\begin{tabular}{*l^l|^c^c^c^c^c^c^c^c^c^c^c^c^c^c^c^c^c^c^c^c^c^c^c^c^c^c^c^c}
CASE & CLASSES  & \datatag{Food101} & \datatag{{CIFAR10}} & \datatag{{CIFAR100}} & \datatag{{Birdsnap}} & \datatag{{SUN397}} & \datatag{{Cars}} & \datatag{{Aircraft}} & \datatag{{VOC2007}} & \datatag{{DTD}} & \datatag{{Pets}} & \datatag{{Cal101}} & \datatag{{Flowers}} & \datatag{{MNIST}} & \datatag{{FER2013}} & \datatag{{STL10}} & \datatag{{EuroSAT}} & \datatag{{RESISC45}} & \datatag{{GTSRB}} & \datatag{{KITTI}} & \datatag{{Country211}} & \datatag{{PCAM}} & \datatag{{UCF101}} & \datatag{{K700}} & \datatag{{CLEVR}} & \datatag{{HM}} & \datatag{{SST}} &  \datatag{{ImageNet}} & \datatag{{AVG}}\\
\midrule
UNICOM&500&45.5 &77.6 &49.8 &27.1 &50.3 &20.1 &21.4 &60.1 &49.9 &80.6 &70.8 &58.0 &94.9 &42.5 &82.6 &91.6 &61.2 &59.4 &60.3 &5.0 &77.4 &45.5 &21.9 &28.8 &44.7 &50.1 &42.1 &52.6\\
UNICOM&1000&51.3 &79.0 &50.7 &28.5 &53.6 &22.1 &22.5 &65.9 &52.0 &83.6 &71.5 &60.6 &95.2 &43.1 &83.7 &92.8 &61.7 &60.6 &63.1 &5.5 &78.6 &46.6 &24.9 &29.3 &45.8 &52.6 &58.4 &54.9\\
UNICOM&2000&52.7 &79.0 &51.3 &29.7 &54.6 &22.8 &24.2 &67.8 &50.7 &84.0 &77.3 &61.0 &95.7 &46.0 &84.4 &93.3 &62.3 &64.2 &65.0 &6.3 &79.4 &46.0 &26.2 &30.4 &47.6 &54.9 &61.5 &56.2\\
UNICOM&4000&54.1 &76.4 &49.7 &35.2 &56.6 &25.3 &28.5 &72.1 &54.1 &82.9 &77.1 &61.6 &95.2 &47.6 &85.3 &94.6 &63.3 &62.4 &66.2 &6.7 &78.7 &48.3 &30.8 &34.6 &50.4 &56.1 &62.8 &57.6\\
UNICOM&8000&54.9 &77.1 &50.5 &35.6 &57.6 &28.7 &28.6 &71.9 &53.3 &82.6 &78.4 &67.8 &95.0 &47.7 &86.3 &94.3 &66.0 &63.5 &66.4 &6.9 &78.1 &52.1 &31.5 &32.8 &50.4 &54.8 &62.4 &58.3\\
UNICOM&20000&56.1 &78.2 &51.1 &38.5 &60.0 &32.6 &32.4 &72.9 &55.1 &84.4 &80.6 &71.2 &95.4 &48.1 &86.6 &95.7 &68.7 &66.3 &65.3 &7.5 &80.8 &55.4 &35.8 &32.8 &49.8 &55.2 &61.5 &59.9\\
UNICOM&40000&57.1 &77.7 &51.8 &38.1 &60.3 &37.3 &35.6 &71.3 &56.9 &83.4 &81.0 &76.8 &95.3 &48.7 &85.4 &96.1 &71.9 &68.9 &68.2 &7.8 &79.3 &56.7 &33.7 &32.8 &50.8 &54.6 &59.5 &60.6\\
UNICOM&80000&57.1 &76.9 &51.5 &38.8 &58.9 &33.9 &35.1 &69.1 &57.4 &81.3 &78.9 &77.4 &95.4 &48.8 &83.1 &95.9 &74.1 &67.3 &67.8 &7.9 &80.3 &55.8 &32.1 &33.8 &49.2 &53.7 &56.4 &59.9\\
UNICOM&160000&56.3 &75.1 &50.1 &37.5 &59.2 &35.1 &35.9 &67.9 &56.5 &79.0 &79.0 &79.5 &96.3 &48.9 &82.3 &96.0 &76.7 &68.4 &70.2 &7.7 &80.2 &57.5 &36.2 &33.0 &51.4 &56.0 &53.6 &60.2\\
UNICOM&320000&53.7 &73.3 &49.8 &34.9 &57.1 &31.7 &37.4 &66.0 &54.8 &74.2 &77.4 &78.4 &96.6 &49.6 &78.3 &95.3 &75.7 &69.4 &69.9 &7.6 &78.7 &56.4 &36.0 &35.6 &50.4 &55.7 &48.5 &59.0\\
UNICOM&500000&48.8 &69.8 &46.2 &28.3 &53.5 &26.9 &36.2 &63.8 &50.3 &67.9 &72.9 &75.0 &97.2 &47.4 &74.3 &94.4 &73.5 &62.3 &73.4 &7.4 &80.0 &52.3 &33.0 &35.6 &49.4 &56.4 &41.6 &56.2\\
\midrule
MLCD&500&55.3 &82.1 &54.3 &41.0 &67.1 &28.1 &35.3 &72.7 &62.3 &87.0 &87.9 &75.3 &97.4 &48.1 &93.1 &94.4 &68.7 &70.7 &64.3 &10.9 &80.8 &60.8 &37.6 &31.6 &50.0 &52.1 &63.2 &61.9\\
MLCD&1000&59.1 &83.2 &59.6 &43.2 &68.1 &30.8 &38.9 &75.2 &64.9 &87.7 &88.5 &77.4 &97.1 &48.5 &94.1 &95.4 &71.3 &72.4 &65.8 &10.6 &79.9 &61.2 &40.6 &34.1 &50.5 &53.7 &67.2 &63.7\\
MLCD&2000&62.0 &84.0 &61.8 &45.4 &69.0 &30.7 &39.7 &77.5 &65.6 &88.0 &88.1 &79.0 &97.0 &49.7 &93.9 &96.7 &74.3 &73.4 &68.8 &10.6 &80.9 &64.6 &41.2 &35.2 &50.6 &54.7 &68.2 &64.8\\
MLCD&4000&65.1 &84.7 &63.5 &47.4 &69.7 &37.2 &41.7 &79.2 &67.6 &88.5 &89.7 &82.6 &97.7 &51.4 &94.4 &97.0 &78.0 &75.7 &70.5 &11.0 &80.5 &67.3 &42.1 &39.8 &50.6 &58.2 &69.9 &66.7\\
MLCD&8000&66.2 &85.1 &64.9 &50.0 &70.4 &42.2 &46.0 &80.2 &68.2 &88.6 &90.2 &84.6 &97.8 &51.5 &94.3 &97.1 &78.9 &78.6 &69.1 &10.8 &81.4 &67.9 &43.7 &37.2 &50.8 &56.4 &69.7 &67.5\\
MLCD&20000&65.9 &84.1 &63.0 &46.5 &68.6 &44.8 &44.9 &81.9 &66.5 &89.9 &89.7 &82.8 &97.5 &51.8 &93.2 &97.1 &78.6 &78.7 &69.5 &10.3 &81.3 &67.1 &42.4 &38.0 &50.2 &56.8 &69.0 &67.0\\
MLCD&40000&65.2 &83.4 &62.4 &48.5 &68.4 &48.3 &46.1 &82.2 &66.6 &90.2 &89.4 &83.1 &97.2 &50.3 &91.9 &96.8 &78.7 &75.2 &70.9 &10.5 &80.0 &66.5 &42.2 &36.6 &52.4 &56.8 &68.8 &67.0\\
MLCD&80000&69.9 &85.6 &65.4 &56.0 &70.7 &57.0 &52.4 &83.2 &68.5 &90.3 &91.2 &88.8 &97.3 &53.3 &93.3 &97.7 &81.2 &79.4 &73.1 &11.5 &79.8 &71.6 &45.1 &38.4 &50.0 &55.5 &70.0 &69.5\\
MLCD&160000&71.3 &86.2 &67.4 &59.2 &71.7 &61.0 &56.1 &84.6 &69.8 &91.3 &91.7 &90.8 &98.0 &53.5 &93.1 &97.7 &83.3 &80.9 &72.7 &11.6 &80.9 &71.9 &46.6 &40.0 &51.8 &55.2 &70.0 &70.7\\
MLCD&320000&72.2 &86.1 &67.4 &60.1 &71.7 &64.8 &56.4 &85.9 &68.7 &90.8 &92.1 &91.7 &98.2 &54.3 &93.2 &98.0 &84.6 &81.9 &74.0 &11.6 &81.3 &73.9 &48.0 &45.2 &49.6 &55.2 &69.7 &71.4\\
MLCD&500000&72.5 &86.3 &68.3 &59.8 &71.7 &65.0 &56.8 &83.1 &69.9 &90.6 &91.7 &92.3 &98.3 &54.7 &93.1 &97.8 &85.2 &82.7 &74.0 &11.5 &81.7 &73.8 &47.1 &44.0 &50.4 &54.5 &68.9 &71.3\\
\bottomrule
\end{tabular}
}

\end{table}

\subsubsection{Linear Probe Evaluation}
In our linear probing analysis, we adhered to the same configuration as CLIP. We utilized the L-BFGS optimization algorithm as implemented in PyTorch, executing it on a GPU with an upper limit of 1000 iterations. We adopted CLIP's parametric binary search protocol to optimize the hyper-parameter $\lambda$, with the optimization process conducted on the validation set. In cases where a dataset lacks a predefined validation set, we manually partition the dataset. This streamlined methodology allowed us to efficiently run tests across all 27 datasets within a few hours. For the final results, the validation set is merged back into the training set for an additional round of training.

\subsubsection{Zero-shot Evaluation}
For the experiments in zero-shot, we use the prompts same as FLIP.
Following CLIP ~\cite{radford2021learning}, we report the mean accuracy per class for FGVC Aircraft, Oxford-IIIT Pets, Caltech-101, and Oxford Flowers 102 datasets. We report the mean of top-1 and top-5 accuracy for Kinetics-700, ROC AUC for Hateful Memes, and 11-point mAP for Pascal VOC 2007 Classification. We report top-1 accuracy for the rest of the datasets.

\subsubsection{Zero-shot Retrieval}
We assess the effectiveness of zero-shot retrieval using two established benchmarks: Flickr30K~\cite{young2014image} and COCO~\cite{lin2014microsoft}, each containing 1K and 5K image-text pairs in their test sets, respectively. In adhering to the procedures outlined in CLIP and FLIP, we derive the image and text embeddings from the relevant encoders, and then execute retrieval by calculating cosine similarities across potential image-text pairs, without prompt being utilized.

\subsubsection{Zero-shot Robustness Evaluation}
In our zero-shot robustness assessment on ImageNet-related sets, we employ the 7 prompts provided by CLIP, with dataset preparation and division adhering to the methods used in OpenCLIP. For ObjectNet, we emulate the approach of CLIP by utilizing class names without any prompt.

\begin{table}
\caption{Comparisons of linear probe performance across 27 different downstream datasets.
The network structure used here is ResNet50 following the BiT~\cite{bigtransfer} and CLIP~\cite{radford2021learning} papers. $\dag$: Results reported in the CLIP paper referring to the linear probe results of BiT~\cite{bigtransfer} with ResNet50 trained on the ImageNet1K dataset. $\ddag$: Results reported in our testing by using the open-sourced BiT~\cite{bigtransfer} ResNet50 model. Different from the baseline model trained with the ground-truth labels, the proposed MLCD models are trained with the automatically clustered class labels.}
\label{tab:imagenet1k}
    \centering
    \resizebox{1.0\linewidth}{!}{
        \tablestyle{0.7mm}{1.1}
        \begin{tabular}{*l^l|^c^c^c^c^c^c^c^c^c^c^c^c^c^c^c^c^c^c^c^c^c^c^c^c^c^c^c^c}
            CASE & CLASSES  & \datatag{Food101} & \datatag{{CIFAR10}} & \datatag{{CIFAR100}} & \datatag{{Birdsnap}} & \datatag{{SUN397}} & \datatag{{Cars}} & \datatag{{Aircraft}} & \datatag{{VOC2007}} & \datatag{{DTD}} & \datatag{{Pets}} & \datatag{{Cal101}} & \datatag{{Flowers}} & \datatag{{MNIST}} & \datatag{{FER2013}} & \datatag{{STL10}} & \datatag{{EuroSAT}} & \datatag{{RESISC45}} & \datatag{{GTSRB}} & \datatag{{KITTI}} & \datatag{{Country211}} & \datatag{{PCAM}} & \datatag{{UCF101}} & \datatag{{K700}} & \datatag{{CLEVR}} & \datatag{{HM}} & \datatag{{SST}} &  \datatag{{ImageNet}} & \datatag{{AVG}}\\
            \midrule

\rowstyle{\color{dt}}RN50$^{\dag}$ &1000&72.5 &91.7 &74.8 &57.7 &61.1 &53.5 &52.5 &83.7 &72.4 &92.3 &91.2 &92.0 &98.4 &56.1 &76.7 &97.4 &85.0 &70.0 &66.0 &12.5 &83.0 &72.3 &47.5 &48.3 &54.1 &55.3 &75.2 &70.1\\
\rowstyle{\color{dt}}RN50$^{\ddag}$ &1000&72.6 &91.5 &74.2 &57.9 &60.1 &51.2 &51.8 &84.1 &70.9 &91.5 &91.5 &91.8 &97.9 &56.1 &77.5 &96.4 &84.7 &73.3 &64.7 &11.4 &83.9 &71.8 &45.4 &44.3 &51.2 &53.4 &75.6 &69.5\\
MLCD&1000&59.1 &83.2 &59.6 &43.2 &68.1 &30.8 &38.9 &75.2 &64.9 &87.7 &88.5 &77.4 &97.1 &48.5 &94.1 &95.4 &71.3 &72.4 &65.8 &10.6 &79.9 &61.2 &40.6 &34.1 &50.5 &53.7 &67.2 &63.7\\
MLCD&160000&71.3 &86.2 &67.4 &59.2 &71.7 &61.0 &56.1 &84.6 &69.8 &91.3 &91.7 &90.8 &98.0 &53.5 &93.1 &97.7 &83.3 &80.9 &72.7 &11.6 &80.9 &71.9 &46.6 &40.0 &51.8 &55.2 &70.0 &70.7\\
MLCD&320000&72.2 &86.1 &67.4 &60.1 &71.7 &64.8 &56.4 &85.9 &68.7 &90.8 &92.1 &91.7 &98.2 &54.3 &93.2 &98.0 &84.6 &81.9 &74.0 &11.6 &81.3 &73.9 &48.0 &45.2 &49.6 &55.2 &69.7 &71.4\\
            \bottomrule
        \end{tabular}
    }
\end{table}

\begin{figure}[t!]
\centering
\includegraphics[width=0.7\textwidth]{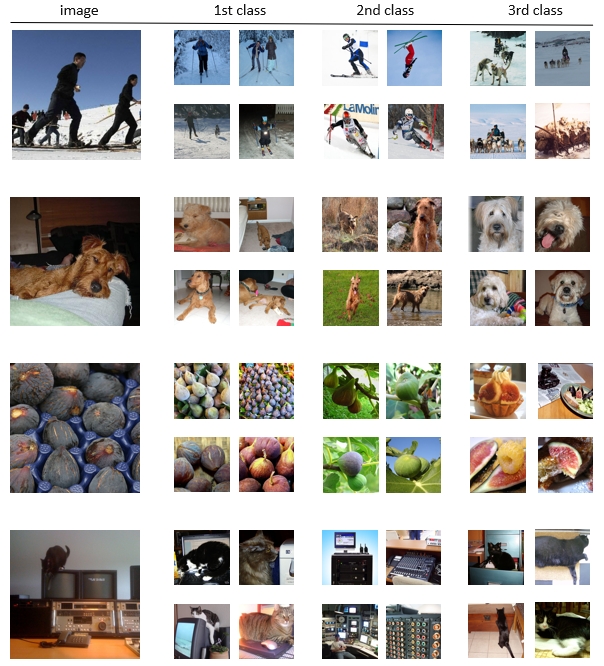}
\caption{Visualization of top 3 labels given to the training samples from the automatically clustered ImageNet dataset. Multiple positive labels show complementary visual signals.}
\label{fig:pruityconflict_in1k}
\end{figure}

\subsubsection{VLM Evaluation}
For VLM evaluation, we tested on the VQAv2~\cite{vlm_vqav2}, GQA~\cite{vlm_gqa}, VizWiz~\cite{vlm_vizwiz}, SQA~\cite{vlm_sqa}, TextVQA~\cite{vlm_text_vqa}, LLaVA-Wild~\cite{liu2024llava_V1_5}, AI2D~\cite{vlm_ai2d}, MathVista~\cite{vlm_mathvista}, HallusionBench
~\cite{vlm_hb}, MMMU~\cite{vlm_mmmu}, cMMMU~\cite{vlm_cmmmu}, MMBench~\cite{vlm_mmbench}, SEED-Bench~\cite{vlm_seedbench} and MME~\cite{vlm_mme} test sets. We used the LMMs-Eval~\cite{lmms_eval2024} tool to evaluate the model. During training, we aligned the hyper-parameters with LLaVA-1.5, using the same pre-training and instruction fine-tuning data as LLaVA-1.5. We also utilized DeepSpeed Zero3\cite{rajbhandari2020zero} to accelerate the training process.

\subsection{Multi-label Learning on ImageNet}

In Tab.~\ref{tab:unicom_mlcd}, we compare the proposed multi-label cluster discrimination and the single-label cluster discrimination (UNICOM~\cite{an2023unicom}) on ImageNet with the clustered class number ranging from 0.5K to 0.5M. The clustering step is conducted by using the features predicted by the CLIP model (\ie, ViT-L/14). In the discrimination step, both UNICOM and MLCD employ the negative class center sampling with a ratio of $0.1$, and the positive number for MLCD is set as $8$. As we can see, the proposed multi-label learning significantly surpasses UNICOM and achieves the best performance of $71.4\%$ when the class number is 320K. In Fig.~\ref{fig:pruityconflict_in1k}, we visualize the top three labels for our training samples. 
When training with multiple labels, our method can
learn complementary visual signals (\eg, different activities in the snow, different breeds of dogs, different locations of figs, and different objects in the room) to improve visual representation learning. 

In Tab.~\ref{tab:imagenet1k}, we compare the performance between models trained with the ground-truth class labels and the automatically clustered class labels. As we can see,
2nd and 3rd rows demonstrate the performance gap between models trained by the ground-truth 1K classes and the automatically clustered 1K classes. Row 4 and 5 indicate that with a significant increase in class numbers (\eg, high purity within each cluster), the results significantly improve. Although the performance of the proposed MLCD on ImageNet does not surpass the supervised case ($69.7\%$ vs. $75.6\%$), the proposed method demonstrates superior feature representation learning capacity across the different datasets ($71.4\%$ vs. $69.5\%$).

\subsection{Acknowledgment}

We would like to thank Tongliang Liu, Jia Guo, and Jing Yang for their insightful discussions on the experimental design of this paper. We thank Bin Qin, Lan Wu, Haiqiang Jiang, and Yuling Wu for their help with the downloading and organization of all the web datasets. We also thank Yin Xie for help with the VLM experiment.

\end{document}